\documentclass[acmlarge]{acmart}
\AtBeginDocument{%
  }

\setcopyright{cc}
\setcctype{by}
\acmJournal{IMWUT}
\acmYear{2025} \acmVolume{9} \acmNumber{4} \acmArticle{193} \acmMonth{12} \acmPrice{}\acmDOI{10.1145/3770709}

\usepackage{titlesec}
\usepackage{multirow}
\usepackage{pifont}
\usepackage{array,tabularx,booktabs,graphicx}

\usepackage{amssymb}
\usepackage{mathtools}
\usepackage{float}
\usepackage{color}
\usepackage{algorithm}
\usepackage{algpseudocode}
\titleformat{\subsubsection}[runin]{\bfseries}{\thesubsubsection}{1em}{}[.\\]

\begin{document}

\title{Gestura: A LVLM-Powered System Bridging Motion and Semantics for Real-Time Free-Form Gesture Understanding}

\author{Zhuoming Li}
\authornote{Both authors contributed equally to this research.}
\author{Aitong Liu}
\authornotemark[1]
\affiliation{%
  \institution{Institute of Artificial Intelligence (TeleAI) of China Telecom}
  \country{China}
}

\author{Mengxi Jia}
\authornote{Corresponding authors}
\affiliation{%
  \institution{Institute of Artificial Intelligence (TeleAI) of China Telecom}
  \city{Shanghai}
  \country{China}
}

\author{Yubo Lu}
\affiliation{%
  \institution{Institute of Artificial Intelligence (TeleAI) of China Telecom}
  \city{Shanghai}
  \country{China}
}

\author{Tengxiang Zhang}
\affiliation{%
  \institution{Goertek Inc}
  \city{Beijing}
  \country{China}
}

\author{Changzhi Sun}
\affiliation{%
  \institution{Institute of Artificial Intelligence (TeleAI) of China Telecom}
  \city{Shanghai}
  \country{China}
}

\author{Dell Zhang$^{\#}$}
\affiliation{%
  \institution{Institute of Artificial Intelligence (TeleAI) of China Telecom}
  \city{Shanghai}
  \country{China}
}
\thanks{$^{\#}$ Project Lead}

\author{Xuelong Li}
\authornotemark[2] 
\affiliation{%
  \institution{Institute of Artificial Intelligence (TeleAI) of China Telecom}
  \city{Shanghai}
  \country{China}
}

\renewcommand{\shortauthors}{Trovato et al.}

\begin{abstract}
Free-form gesture understanding is highly appealing for human-computer interaction, as it liberates users from the constraints of predefined gesture categories. However, the sole existing solution—GestureGPT—suffers from limited recognition accuracy and slow response times. In this paper, we propose Gestura, an end-to-end system for free-form gesture understanding.
Gestura harnesses a pre-trained Large Vision-Language Model (LVLM) to align the highly dynamic and diverse patterns of free-form gestures with high-level semantic concepts.
To better capture subtle hand movements across different styles, we introduce a Landmark Processing Module that compensate for LVLMs' lack of fine-grained domain knowledge by embedding anatomical hand priors.
Further, a Chain-of-Thought (CoT) reasoning strategy enables step-by-step semantic inference, transforming shallow knowledge into deep semantic understanding and significantly enhancing the model’s ability to interpret ambiguous or unconventional gestures.
Together, these components allow Gestura to achieve robust and adaptable free-form gesture comprehension. Additionally, we have developed the first open-source dataset for free-form gesture intention reasoning and understanding with over 300,000 annotated QA pairs. Experimental results show that Gestura achieves the accuracy of 84.73\% (closed-set) / 64.14\% (open-set) in the exocentric (third-person) setting and 66.14\% (closed-set) / 21.71\% (open-set) in the egocentric (first-person) setting, achieving approximately \textbf{20\% and 40\% higher accuracy} on closed-set and open-set tasks, respectively, compared to GestureGPT. Moreover, Gestura achieves over a \textbf{100× speedup} in response time (1.6 seconds vs. 227 seconds) on an 8B-sized model deployed on a single NVIDIA A100 40GB GPU, and has been validated through real-device experiments with an edge–cloud collaborative setup, bringing free-form gesture understanding markedly closer to practical, real-world deployment. 
Both the dataset and code about the project can be accessed at https://evans-lx.github.io/Gestura/.
\end{abstract}




\maketitle

\section{Introduction}
Gesture-based interaction serves as a critical interface for wearable devices such as smart glasses \cite{10.1145/3675094.3678992,10.1145/3463509,10.1145/3610910}, enabling intuitive human-device communication. However, real-world deployment faces a significant challenge: spontaneous gesture understanding in open-world environments. As these devices become more pervasive in daily life, users expect to interact naturally—using spontaneous, personalized gestures instead of memorizing predefined commands. This shift demands gesture recognition systems that are not only accurate but also adaptable to individual styles, cultural variations, and situational context. Unlike controlled settings, users exhibit diverse, context-driven gestures that carry implicit intentions, demanding systems capable of parsing both motion patterns and latent semantics. Traditional approaches relying on predefined gesture libraries or specialized hardware (\emph{e.g.}, gloves, depth sensors) fail to address this complexity, creating barriers to natural interaction and scalability.


Current gesture recognition methods fall into two categories. \emph{i.e.}, 1) Conventional deep learning methods (\emph{e.g.}, CNNs, LSTMs) excel at classifying fixed gesture vocabularies, but lack semantic reasoning to infer intentions from novel motions. 2) Emerging Large Language Models (LLMs) such as GestureGPT \cite{zengGestureGPTZeroshotInteractive2024}, though attempting open-world interpretation, face dual constraints: Firstly, they depend on intensive computational resources and introduce significant latency, which hinders real-world deployment; more critically, existing LLMs lack understanding of fine-grained hand dynamic, which fundamentally limits their capacity to capture nuanced gestural semantics. 
This presents a crucial research challenge: how to adaptively integrate fine-grained motion cues with general reasoning capabilities to effectively bridge low-level motion tracking and high-level intent inference. This integration constitutes a critical pathway for free-form gesture understanding.

To address this challenge, we turn to Large Vision-Language Models (LVLMs)—an emerging extension of LLMs that incorporate both visual and textual modalities. While early LLMs were limited to text-only inputs, recent advances have enabled them to process images and videos through aligned visual encoders and MLP projector. These LVLMs retain the strong reasoning capabilities of LLMs, while gaining the capacity to perceive complex visual contexts.

This paper explores a novel synergy: leveraging robust hand landmark priors to "teach" a LVLM to resolve ambiguities in free-form gesture understanding. Our key insight is that while traditional hand keypoint detection models offer geometrically accurate but semantically shallow representations, they can provide crucial contextual information to LVLM that lack domain-specific knowledge but possess strong reasoning abilities. 
Our framework enables kinematic precision to inform contextual semantics through innovative cross-modal interaction.

Specifically, we propose a hierarchical framework centered around \textbf{two tightly integrated components}: (1) a Landmark Processing Module, which incorporates hand keypoint information extracted from MediaPipe \cite{lugaresi2019mediapipeframeworkbuildingperception} to enrich visual features with anatomical structure and spatial cues—crucial for distinguishing subtle gesture variations; (2) a LVLM backbone that employs a dual-stream encoder and a large language model to align dynamic visual features with high-level semantic concepts, enabling robust intent inference beyond superficial motion patterns; and \textbf{a two-stage training paradigm}:  
First, Gestura learns general visual-semantic mappings through a multi-view semantic enhancement strategy to activate the model’s potential for free-form generalization. 
Subsequently, in Stage 2, Gestura leverages a well-trained landmark processing module to transmit anatomical and spatial contextual signals to the LVLM backbone. Meanwhile, it internalizes advanced reasoning capabilities through Chain-of-Thought (CoT) tuning. This enables Gestura to push the model's reasoning limits in open-world scenarios and achieve superior generalization.

To evaluate the effectiveness of our model, we conducted comprehensive experiments across both benchmark and real-world scenarios.
Specifically, Gestura achieves a top-1 intent accuracy of 84.73\% in the exocentric view and 66.14\% in the egocentric view, substantially outperforming the previous state-of-the-art, GestureGPT, which achieves 72.08\% (closed-set) / 44.46\% (open-set) and 40.07\% (closed-set) / 17.38\% (open-set) under the corresponding settings.
This highlights a significant advancement in accurate gesture intent interpretation, particularly under the challenging egocentric perspective.
Moreover, on gesture description tasks, Gestura also demonstrates a substantial improvement: achieving BLEU-4 scores of 49.83 (exocentric, closed-set) and 43.67 (egocentric, closed-set), compared to GestureGPT’s 15.05 and 10.75, respectively—representing over 3× higher description quality. Even under open-set conditions, Gestura maintains strong performance, with BLEU-4 scores of 14.17 (exocentric) and 9.93 (egocentric), further validating its robustness in free-form gesture understanding. The relatively lower open-set accuracy in the egocentric setting can be attributed to the smaller scale of egocentric data compared to exocentric data in our dataset, which limits the model’s ability to generalize to novel gestures from first-person perspectives.
Beyond benchmarks, we deployed Gestura in practical egocentric smart home settings, where it reached a top-5 accuracy of 69.23\% across eight free-form intent categories, proving its robustness to gesture variability and its suitability for real-time, context-aware human-computer interaction. Notably, the average response time in real-device experiment settings is 7.83 seconds, including communication and TTS latency, while pure model inference takes only 1.6 seconds—significantly faster than prior models like GestureGPT (227 seconds)—making Gestura highly efficient for interactive applications.

Our contributions can be summarized as follows:




\begin{enumerate}
    \item \textbf{A novel dataset for free-form gesture intent understanding:} We present a pioneering dataset that redefines gesture data through a question-answering (QA) paradigm, specifically designed for free-form, user-defined gestures. Unlike traditional datasets constrained by fixed vocabularies, ours captures the diversity of real-world gestures by including rich annotations at three levels: \textit{action description}, \textit{gesture meaning}, and \textit{contextual intent}. This enables robust benchmarking for intent reasoning under open-world conditions.
    
    \item \textbf{A LVLM-based inference architecture tailored for free-form gesture comprehension:} We extend a LVLM to support fine-grained gesture reasoning by aligning motion patterns with latent semantics. Our architecture bridges low-level visual encoding and high-level intention inference, allowing it to generalize across unseen gestures. Crucially, our system achieves this while maintaining real-time performance (\textit{7.83 s} per gesture), making it possible for deployment in wearable and edge-computing scenarios.
    
    \item \textbf{Semantic grounding through anatomical priors:} To better differentiate between various hand gestures, we utilize Mediapipe as an auxiliary tool by integrating the gesture ground truth detected by Mediapipe with video data. This approach proves beneficial in distinguishing similar gestures, particularly those with visually similar features. In essence, our idea is to leverage the prior knowledge of a specialized gesture recognition model to enhance the understanding of gestures by our language model (LM). This fusion enables the LM to interpret gestures more accurately, facilitating the model to better distinguish fine-grained differences between different gestures.
\end{enumerate}

\begin{figure}[h]
  \centering
  \includegraphics[width=\linewidth]{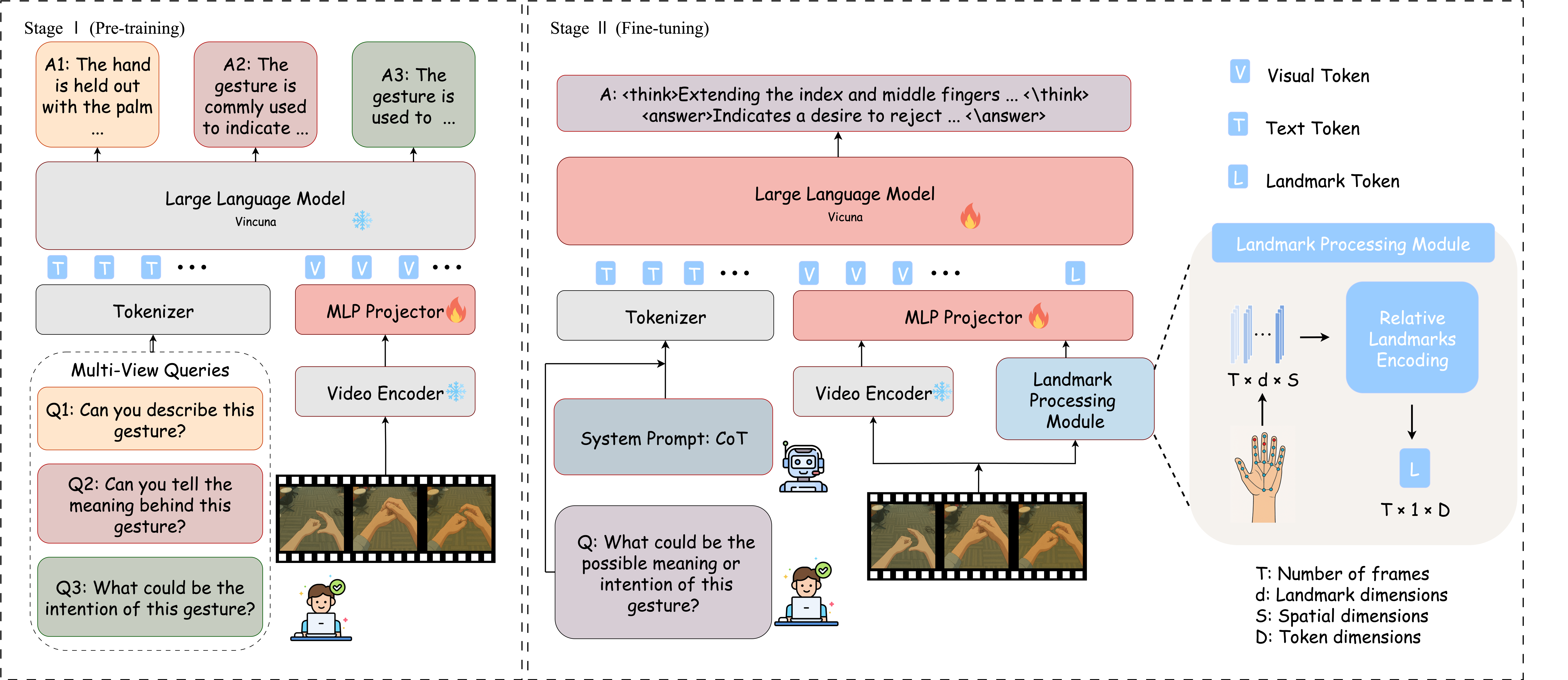}
  \caption{\textbf{Overview of the proposed framework of Gestura. Gestura introduces a  hierarchical framework with two-phase training for free-form gesture understanding.} First, the pre-training stage activates the model's potential for free-form generalization by using a multi-view semantic enhancement strategy. 
  In Stage 2, Gestura leverages well-trained landmark processing module to deliver anatomical and spatial contextual signals to the LVLM backbone, while internalizing advanced reasoning capabilities through Chain-of-Thought (CoT) tuning. This dual mechanism expands the model’s reasoning boundaries, enabling superior generalization in free-form gesture understanding.}

  \Description{A woman and a girl in white dresses sit in an open car.}
\end{figure}
\section{Related Work}
Gesture recognition has garnered significant attention in both human-computer interaction (HCI) and human-robot interaction (HRI) domains \cite{bandiniAnalysisHandsEgocentric2022,qiComputerVisionbasedHand2024}. Recent advances in hand gesture recognition, particularly in dynamic settings, have leveraged different techniques and data modalities such as RGB images, depth sensors, and 3D skeleton-based data \cite{linardakisSurveyHandGesture2025,miahDynamicHandGesture2023}. The use of 3D hand skeleton data, particularly hand landmarks, has become an important aspect of gesture recognition due to its robustness and ability to capture rich spatial-temporal features. Many systems rely on spatial-temporal attention mechanisms to model dependencies between hand joints, making them effective for real-time recognition tasks.

In recent years, significant progress has been made in dynamic hand gesture recognition (DHGR) using deep learning models such as convolutional neural networks (CNNs) \cite{sunOptimizingGestureRecognition2024,almudawiInnovativeHealthcareSolutions2024}, graph-based neural networks, and long short-term memory (LSTM) networks \cite{rastgooMultimodalZeroshotDynamic2024}. For instance, graph-based architectures have been proposed to better handle the spatial and temporal dependencies between joints for improved performance on skeleton-based hand gesture datasets, such as SHREC’17 \cite{Rodol2017SHREC1}, achieving high accuracy in recognizing complex gestures. These approaches focus on extracting spatial-temporal attention features to capture joint dependencies, allowing for real-time and efficient gesture recognition.

More recently, the emergence of large-scale pre-trained models—particularly transformer-based architectures has further propelled the development of HCI and gesture recognition systems \cite{khaokaewMAPLEMobileApp2024,kingSashaCreativeGoalOriented2024,sunMultimodalDailyLifeLogging2024,xiaoChatCamEmbracingLLMs2024}. These models, originally popularized in natural language processing, have demonstrated remarkable capacity in capturing long-range dependencies and multi-level abstractions, making them highly suitable for tasks involving sequential and multi-dimensional inputs, such as gesture sequences. Vision Transformers \cite{DBLP:journals/corr/abs-2010-11929}, for example, have been adapted to process RGB frames or skeletal heatmaps \cite{liuTagSleep3DRFbased3D2024}, showing competitive performance in modeling global gesture contexts.

At the same time, the growing integration of wearable devices and ubiquitous computing technologies has significantly enhanced gesture recognition capabilities \cite{huIOTeethIntraOralTeeth2024,shimonExploringUnimanualEar2024,wangUFaceYourSmartphone2024}. Wearables such as smart glasses, motion sensors, and inertial measurement units (IMUs) provide continuous, fine-grained motion data that complements traditional visual modalities \cite{huangSpeciFingersFingerIdentification2024}. These devices enable gesture recognition systems to operate in unconstrained environments with improved responsiveness and contextual awareness. When combined with transformer-based models, the rich multimodal signals from wearable platforms can be effectively aligned and interpreted, paving the way for more natural, adaptive, and personalized interaction experiences across a wide range of real-world applications.

However, while significant advancements have been made in recognizing predefined gestures, most existing systems require users to learn and perform gestures from a fixed set, which can result in a less natural interaction experience. Recent work, such as GestureGPT, proposes a solution for free-form gesture recognition by enabling automatic understanding of spontaneous gestures. This approach eliminates the need for users to learn predefined gestures, allowing for intuitive interaction without explicit training. GestureGPT utilizes LLM to interpret hand gestures from natural language descriptions, mapping them directly to interface functions, thus overcoming the rigid nature of traditional gesture recognition systems. The framework achieves strong performance in zero-shot gesture recognition.

Moreover, a recent exploration into Zero-Shot Learning (ZSL) for dynamic hand gesture recognition proposes a novel multi-modal ZSL framework \cite{rastgooMultimodalZeroshotDynamic2024}. This method combines deep learning based features with skeleton-based representations to perform gesture recognition without annotated data. By leveraging transformer models and BERT-based semantic mapping, the system achieves significant performance improvements in recognizing hand gestures without requiring specific gesture annotations. This approach is particularly relevant when handling open-world tasks where the gesture set is not fixed, and new gesture categories need to be recognized on the fly.

In the realm of open-world gesture recognition, methods have also been proposed to tackle the challenges of catastrophic forgetting and incremental learning \cite{shenOpenWorldGestureRecognition2024}. The work on Data-Free Class-Incremental Learning \cite{aichDataFreeClassIncrementalHand2023} introduces a method for dynamic hand gesture recognition that doesn’t require access to previously seen data, making it suitable for real-world applications where privacy constraints prevent data storage. By employing a boundary-aware prototypical sampling mechanism, the approach improves model inversion and recognition performance without compromising on accuracy or computational efficiency. This method demonstrates its effectiveness in 3D skeleton gesture recognition by addressing both the challenges of incrementally adding new gestures and avoiding performance degradation on old gestures.

Additionally, the synthesis of stylized gestures for human-robot interaction has gained attention, with systems like GestureDiffuCLIP \cite{aoGestureDiffuCLIPGestureDiffusion2023} offering a flexible control mechanism for generating gestures with varying levels of style. By using the CLIP model for style conditioning and a latent diffusion model for gesture generation, this framework allows for high-quality stylized gestures tailored to different interactive scenarios. While this research focuses on synthesizing co-speech gestures, its underlying mechanism for gesture generation and style control could be beneficial in the context of interactive gesture recognition systems, where natural and intuitive user interactions are crucial.

Together, these studies contribute to the growing body of work on dynamic and free-form gesture recognition. They highlight the challenges in achieving real-time performance, maintaining model generalizability, and adapting to open-world environments \cite{higgerOpenWorldHumanRobotInteraction}. Our work contributes a unified framework that leverages structured hand landmark priors to enhance the semantic reasoning capabilities of LVLM, achieving fine-grained and interpretable recognition of free-form gestures in open-world scenarios through hierarchical alignment and progressive training.

\section{Approach}
\subsection{Overview}
Our framework introduces a hierarchical approach for gesture understanding, combining landmark-enhanced visual processing, cross-modal alignment, and progressive training strategies to bridge gesture dynamics with semantic intent.

Landmark Processing Module augments raw video features with structural hand keypoints extracted via MediaPipe. By encoding spatial relationships  between 21 hand landmarks, it enriches visual representations with geometric cues, enabling finer distinction between subtle gesture variations.

LVLM integrates a dual-stream video encoder (capturing static poses and dynamic motions) with a MLP projector that maps visual features into a shared language space. A LLM then understand these features, leveraging its reasoning capabilities to infer gesture meanings beyond surface-level motions. 

Training Pipeline follows a two-stage paradigm:

\textbf{Stage 1.}In the initial pre-training stage, we freeze the parameters of the video encoder and the LLM and introduce a trainable MLP projector that serves as a bridge, aligning video features with textual captions. The reason of freezing the parameters of the video encoder is to preserve its spatiotemporal pattern recognition capabilities. This design is motivated by a deep understanding of gestural motion regularities—such as the rhythm of waving or the trajectory of finger flexion—which are fundamental and transferable across tasks. Direct fine-tuning could risk disrupting these fine-grained motion representations. Holding the LLM weights fixed prevent premature drift, so the model first learns a clean alignment between vision tokens and the LLM's embedding space. This stabilises autoregressive convergence and lays a solid cross-modal foundation\cite{10.5555/3666122.3667638, lin2024videollavalearningunitedvisual}. The relevant ablation experiments could be seen in Appendix ~\ref{app:B}.

Specifically, the MLP is trained to map different semantic expressions of the same gesture sample into a shared embedding space while pushing apart unrelated gestures. To enable such fine-grained alignment, we adopt a multi-view semantic enhancement strategy: (1) Motion Description (e.g., “palm extended”), (2) Gesture Semantics (e.g., “inviting”), and (3) Intent Inference (e.g., “beckon”). This hierarchical structure guides the model in learning layered representations and decoupling low-level motion patterns from high-level semantics. 

\textbf{Stage 2.}Once the model has acquired a baseline capacity for semantic alignment, the second stage focuses on resolving ambiguity between gestures with similar meanings or similar appearances in open-world scenarios. Here, MediaPipe landmark features are introduced and integrated via an anatomically constrained attention mechanism. This dynamic feature selection allows the model to distinguish between confusable gestures based on subtle joint angle variations and coherent motions.

Furthermore, the incorporation of Chain-of-Thought (CoT) tuning breaks the limitations of traditional end-to-end models by explicitly decomposing the reasoning process into a structured sequence: gesture description$\rightarrow $semantic analogy $\rightarrow$ intent hypothesis $\rightarrow$ final decision. This enforces logical traceability and encourages the model to form robust associations rather than overfitting to surface-level patterns. As a result, the model achieves improved accuracy in open-ended scenarios and gains interpretability through intermediate reasoning steps.

This two-stage framework mirrors the biological plausibility of human cognition. Just as humans build a repertoire of motion patterns through repeated observation (analogous to Stage 1: pattern recognition), and then infer intent by incorporating contextual cues (analogous to Stage 2: semantic reasoning), our system enables a qualitative leap from motion decoding to cognitive-level interpretation. In this process, MediaPipe landmarks act as a form of “proprioception”—providing the model with physical constraints akin to the human sense of joint position—while CoT-based reasoning simulates expert-like inference grounded in domain knowledge (e.g., anatomical constraints on elbow movement), enabling more accurate intent disambiguation.

\subsection{Landmark Processing Module}
\begin{figure}[h]
  \centering
  \includegraphics[width=0.8\linewidth]{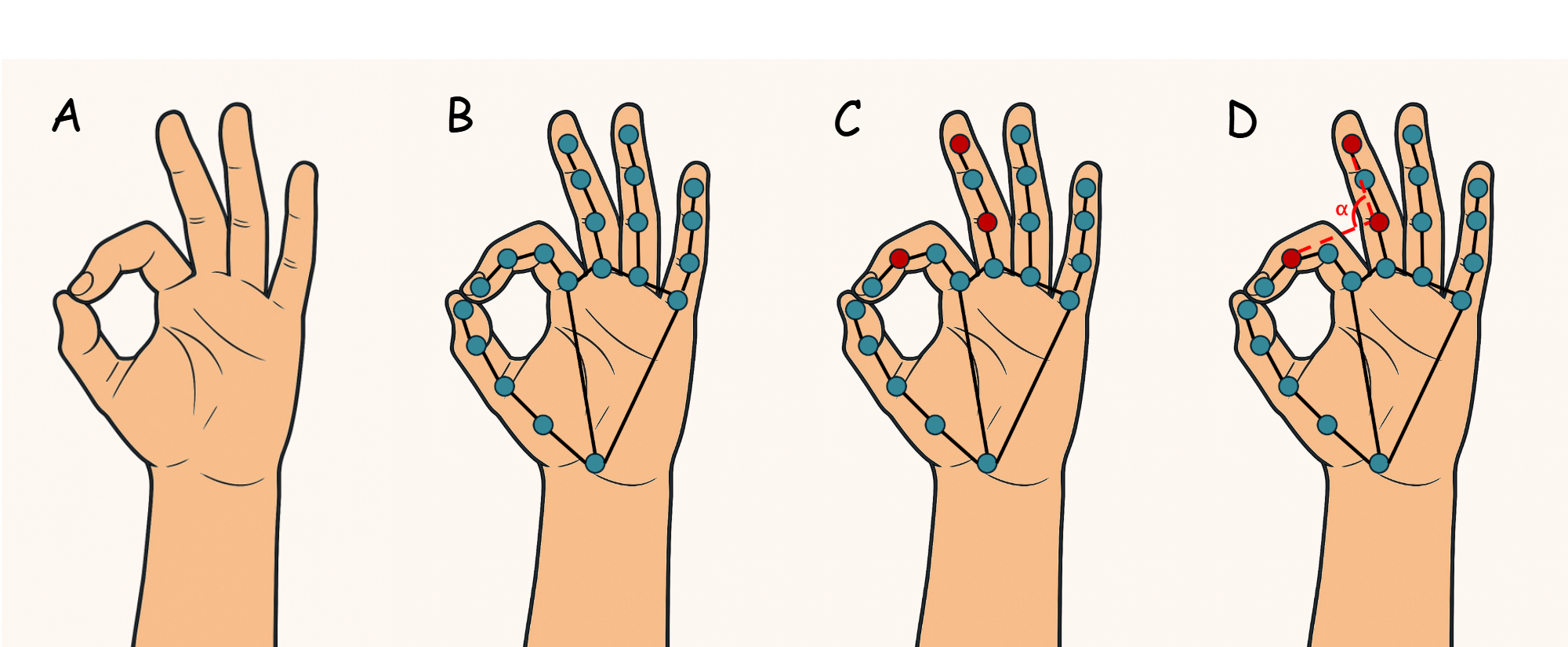}
  \caption{\textbf{The process of deriving encoded features from hand video data. [R1, R4] A. Raw RGB frame → B. 21 MediaPipe landmarks → C. One landmark triplet (red) case from the 1,330 possible → D. Measure the enclosed angle.} Blue circles mark the 21 hand landmarks detected by \textsc{MediaPipe}. One triplet of landmarks (red) is chosen out of the $\binom{21}{3}=1{,}330$ possible combinations each time to form two vectors whose enclosed angle~$\alpha$ is measured. The cosine of this angle constitutes a single element of the final feature vector.}
\end{figure}
Considering that merely encoding videos through a video encoder is insufficient for the LLM to effectively distinguish the subtle differences in vision features between similar gestures, we have added ground information extracted by Mediapipe to the original vision features. The ground information here refers to the 21 key points of the hand extracted by Mediapipe, with each key point represented by its x, y, and z spatial positions.  

For each gesture frame, $n$ keypoints $p_1, p_2, \dots, p_n$ are extracted, where $p_i = (x_i, y_i, z_i)$. Derived features include pairwise distances:
\[
d_{ij} = \| p_i - p_j \|_2 = \sqrt{(x_i - x_j)^2 + (y_i - y_j)^2 + (z_i - z_j)^2}
\]
and angles between keypoints:
\[
\theta_{ijk} = \cos^{-1}\left( \frac{(p_j - p_i) \cdot (p_k - p_i)}{d_{ij}d_{ik}} \right)
\]
which is treated as the ground truth feature.

To prevent the scale differences between videos from affecting the ground information, we chose to use the cosine values formed by every three points as the final ground feature. Among the 1,330 possible combinations of selecting three points from the 21 key points, we used only 1,024 combinations for feature extraction to facilitate subsequent concatenation with vision features. This is because selecting any three points from the hand keypoints results in a total of 1,330 combinations. Considering that the hidden dimension of the vision token is 1,024, and for ease of concatenation, we chose the first 1,024 combinations to compute the cosine similarity values, which are used as the hidden dimension for the landmark tokens. To avoid distortion of the encoded features, we refrained from selecting all combinations and mapping them to 1,024. Since 1,024 combinations cover nearly all possibilities, we selected the first 1,024 combinations for the hands in all frames, ensuring that the features remain intact and consistent.

After adding the ground information, the number of feature tokens corresponding to each frame increased from 257 to 258, enriching the feature representation of gesture videos. The more comprehensive feature representation enables the model to better distinguish fine-grained differences between different gestures.

\subsection{Large Vision Language Model}
\begin{figure}[h]
  \centering
  \includegraphics[width=\linewidth]{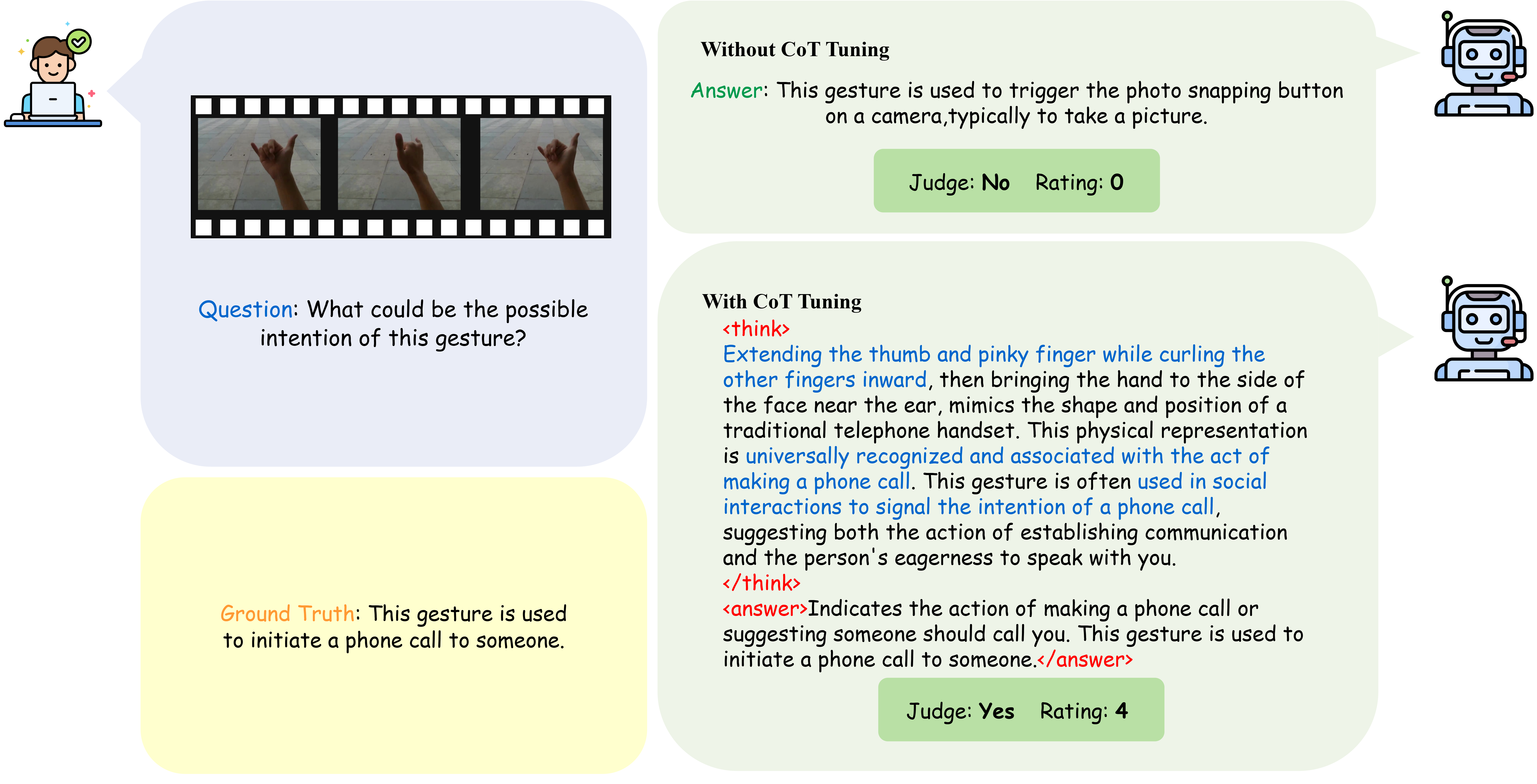}
  \caption{\textbf{Score rating and comparison of answers with and without CoT tuning.} Certain portions of the generated output are highlighted to underscore their critical role in the inference process leading to the final answer.}
  \Description{A woman and a girl in white dresses sit in an open car.}
\end{figure}
To align gesture videos with their textual descriptions, we employ a video encoder that captures both static hand poses and dynamic motion patterns, ensuring comprehensive spatial-temporal representations. These gesture features are then projected into a shared semantic space using an MLP projector. In parallel, gesture-related text is tokenized into structured tokens. Both visual and textual embeddings are fed into an LLM, leveraging its reasoning and contextual understanding to bridge motion and semantics—allowing the model to infer gesture intent beyond surface-level motion.






\subsection{Training Pipeline}

\subsubsection{STAGE 1: Multi-view semantic enhancement}
In the pre-training stage, we align gesture video features with textual representations in a unified feature space. The video encoder extracts both static spatial features and dynamic motion patterns from gesture videos, which are then transformed by a trainable MLP projector. The video encoder remains frozen, ensuring that its pre-trained weights are preserved, while the MLP projector adapts the video features optimally.

The raw video feature vector from the frozen video encoder is denoted as 
$\mathbf{v} \in \mathbb{R}^{d_v}$.
The MLP projector transforms this into the unified feature space as:
\begin{equation}
\mathbf{z} = \underset{d_z \times d_h}{W_3} \cdot \text{GELU}\Big( \underset{d_h \times d_h}{W_2} \cdot \text{GELU}\big( \underset{d_h \times d_v}{W_1} \mathbf{v} + \mathbf{b}_1 \big) + \mathbf{b}_2 \Big) + \mathbf{b}_3
\end{equation}
where:
\begin{itemize}
    \item $\mathbf{v} \in \mathbb{R}^{d_v}$: Input video features
    \item $W_1 \in \mathbb{R}^{d_h \times d_v}, \mathbf{b}_1 \in \mathbb{R}^{d_h}$: First layer parameters
    \item $W_2 \in \mathbb{R}^{d_h \times d_h}, \mathbf{b}_2 \in \mathbb{R}^{d_h}$: Hidden layer parameters 
    \item $W_3 \in \mathbb{R}^{d_z \times d_h}, \mathbf{b}_3 \in \mathbb{R}^{d_z}$: Output layer parameters
    \item $\text{GELU}$: Gaussian Error Linear Unit activation
    \item $\mathbf{z} \in \mathbb{R}^{d_z}$: Projected output features
\end{itemize}

To improve semantic alignment, we employ a multi-view semantic enhancement strategy, analyzing gestures from three distinct perspectives before passing them into the MLP projector. This process refines the representation by capturing nuanced aspects of gesture execution. The resulting video tokens, together with text tokens obtained via a tokenizer, are then fed into a frozen LLM. By maintaining a shared latent space and freezing the LLM parameters, we preserve its strong language reasoning capabilities while enabling the MLP projector to bridge the modality gap between vision and text. Stage 1 relies on a curated subset of approximately 400
000 question–answer pairs paired with 160 000 videos to establish robust descriptive and semantic grounding

The resulting video tokens, combined with text tokens from a tokenizer, are input into a frozen LLM. Freezing the LLM preserves its language reasoning capabilities, while the MLP projector bridges the vision-text modality gap.

\begin{algorithm}[htbp]
\caption{Multi-view Semantic Enhancement (Pre-training)}
\label{alg:stage1}
\begin{algorithmic}[1]
\Require video dataset $\mathcal{V}$ with three dimension annotation, pretrained VideoEncoder $f_{\mathrm{video}}$, LLM, epochs $E_{1}$, batch size $B$
\Ensure trained MLP projector $P_{\mathrm{proj}}$
\State initialize $P_{\mathrm{proj}}$ parameters $\{W_{1},b_{1},W_{2},b_{2},W_{3},b_{3}\}$
\State freeze VideoEncoder, freeze LLM
\For{$e = 1$ to $E_{1}$}
  \For{each batch $\{v_{i}\}_{i=1}^{B}$ from $\mathcal{V}_1$}
    \State $v_{\mathrm{raw}} \gets f_{\mathrm{video}}(v_{i})$
    \State $\mathrm{views} \gets \mathrm{generate\_three\_views}(v_{i})$
    \State $\mathrm{z\_views}\gets\emptyset$
    \For{each $view$ in $\mathrm{views}$}
      \State $v_{\mathrm{view}}\gets f_{\mathrm{video}}(view)$
      \State $h_{1}\gets\mathrm{GELU}(W_{1}\,v_{\mathrm{view}} + b_{1})$
      \State $h_{2}\gets\mathrm{GELU}(W_{2}\,h_{1} + b_{2})$
      \State $z_{\mathrm{view}}\gets W_{3}\,h_{2} + b_{3}$
      \State append $z_{\mathrm{view}}$ to $\mathrm{z\_views}$
    \EndFor
    \State $z_{\mathrm{video}}\gets\mathrm{combine}(\mathrm{z\_views})$
    \State $\mathrm{tokens\_text}\gets \mathrm{LLM.tokenize}(t_{i})$
    \State $\mathrm{tokens\_video}\gets \mathrm{map\_to\_embeddings}(z_{\mathrm{video}})$
    \State $\mathrm{logits}\gets \mathrm{LLM.forward}(\mathrm{tokens\_video}\,\|\,\mathrm{tokens\_text})$
    \State $\mathrm{loss}\gets \mathrm{AlignmentLoss}(\mathrm{logits},\mathrm{tokens\_text})$
    \State $\mathrm{loss.backward}()$ 
    \State optimizer.step(); optimizer.zero\_grad()
  \EndFor
\EndFor
\State \Return $P_{\mathrm{proj}}$
\end{algorithmic}
\end{algorithm}

\subsubsection{STAGE 2: Cross-modal interaction with CoT tuning}




During pre-training with individually separated questions, we observed strong performance on closed-world test data but significant drops in accuracy under open-world scenarios. To address this issue, we aim to equip the model with the ability to reason from gesture descriptions to their intended meanings. We modified the system prompt to encourage causal reasoning across gesture, context, and interpretation while fine-tuned the model using the CoT-formed data to cultivate a habit of chain-of-thought style responses similarly to the cold-start phase of DeepSeek-R1\cite{deepseekai2025deepseekr1incentivizingreasoningcapability}. Furthermore, we incorporated hand landmark information to improve the model’s ability to distinguish between different gestures.

With these modifications, the increased accuracy of open-world test scenarios demonstrating significant improvements in generalization and reasoning capabilities. The training details of the second stage are as follows:

Let $H(t)$ be the hand landmark coordinates at time $t$ extracted by MediaPipe, $V(t)$ be the raw video data at time $t$, and $P_\theta$ be the projector network. The ground truth feature encoding is then computed as:

\begin{equation}
\mathbf{G}(t) = f_{\text{rle}}\Big(f_{\text{mp}}(\mathbf{V}(t)\Big)
\end{equation}

where:
\begin{itemize}
\item $f_{\text{mp}}(\cdot)$ is the MediaPipe Landmark Extraction function
\item $f_{\text{rle}}(\cdot)$ is the Relative Landmarks Encoding function
\end{itemize}

\begin{equation}
\mathbf{z}(t) = P_\theta\Big([\mathbf{G}(t); f_{\text{video}}(\mathbf{V}(t))]\Big)
\end{equation}

where:
\begin{itemize}
\item $f_{\text{video}}(\cdot)$ extracts deep video features
\item $[\cdot;\cdot]$ denotes feature concatenation
\item $P_\theta$ is the multi-layer projector network with GELU activations
\end{itemize}


In this stage, the system simultaneously processes two input streams: (1) ground truth features containing annotated gesture information, and (2) raw video features extracted from the visual encoder. These features are combined and transformed through the multi-layer projector to produce modality-aligned embeddings.

In addition to feature augmentation, the fine-tuning stage also modifies the training paradigm by structuring the data used in stage 1 into a Chain-of-Thought (CoT) format to facilitate step-by-step reasoning over gesture intention, yielding a curated set of 110,000 examples. This transformation systematically links gesture video features, fine-grained keypoint information, and the underlying intent in a step-by-step reasoning framework. By integrating structured reasoning chains, the model is encouraged to progressively infer gesture intent rather than relying on surface-level pattern recognition. This procedure is similar to process supervision\cite{10.5555/3666122.3669222}, which further endows the model with enhanced reasoning capabilities. 

Furthermore, the system prompt used in the large language model (LLM) is carefully redesigned to stimulate its reasoning capabilities\cite{10.5555/3600270.3602070}, guiding it to establish logical connections between gesture actions, contextual cues, and inferred meanings. This refined approach not only improves accuracy in distinguishing similar gestures but also enhances the model’s interpretability, making it better suited for real-world applications where precise gesture understanding is crucial.

\begin{algorithm}[htbp]
\caption{Cross-modal Interaction with CoT Tuning (Fine-tuning)}
\label{alg:stage2}
\begin{algorithmic}[1]
\Require video dataset $\mathcal{V}$ with CoT, pretrained VideoEncoder $f_{\mathrm{video}}$, LLM, epochs $E_{2}$, batch size $B$
\Ensure trained ground-MLP projector $P_{\mathrm{ground}}$
\State initialize $P_{\mathrm{ground}}$ parameters $\theta$
\State freeze VideoEncoder, unfreeze LLM
\State $mp\_extractor \gets \mathrm{MediaPipeLandmarkExtractor}()$
\State $rle\_encoder \gets \mathrm{RelativeLandmarksEncoder}()$
\For{$e = 1$ to $E_{2}$}
  \For{each batch $\{v_{i}\}_{i=1}^{B}$ from $\mathcal{V}_2$}
    \State $H_{i}\gets mp\_extractor.extract(v_{i})$
    \State $G_{i}\gets rle\_encoder.encode(H_{i})$
    \State $V_{\mathrm{feat}}\gets f_{\mathrm{video}}(v_{i})$
    \State $C_{i}\gets [G_{i}\,\|\,V_{\mathrm{feat}}]$
    \State $h_{1}\gets\mathrm{GELU}(\theta\cdot W_{1}\,C_{i} + \theta\cdot b_{1})$
    \State $h_{2}\gets\mathrm{GELU}(\theta\cdot W_{2}\,h_{1} + \theta\cdot b_{2})$
    \State $z_{\mathrm{fine}}\gets \theta\cdot W_{3}\,h_{2} + \theta\cdot b_{3}$
    \State $cot\_prompt \gets \mathrm{build\_CoT\_prompt}(z_{\mathrm{fine}})$
    \State $\mathrm{tokens\_prompt}\gets \mathrm{LLM.tokenize}(cot\_prompt)$
    \State $\mathrm{tokens\_target}\gets \mathrm{LLM.tokenize}(\mathrm{intent}_{i})$
    \State $\mathrm{output}\gets \mathrm{LLM.generate}(\mathrm{tokens\_prompt})$
    \State $\mathrm{loss}\gets \mathrm{CrossEntropy}(\mathrm{output},\mathrm{tokens\_target})$
    \State $\mathrm{loss.backward}()$ 
    \State optimizer.step(); optimizer.zero\_grad()
  \EndFor
\EndFor
\State \Return $P_{\mathrm{ground}}$
\end{algorithmic}
\end{algorithm}
\section{Experiment}

\subsection{Training Details}
In our work, we adopt the Vicuna-7b v1.5 model as the language backbone. Vicuna is an instruction-following large language model fine-tuned from Llama 2 using user-shared conversations collected from ShareGPT. In the training process, we uniformly sample 8 frames from each video, and each frame undergoes preprocessing where it is resized and cropped to a size of 224 $\times$ 224. In the first stage, we train for one epoch with a batch size of 256, keeping both the video encoder and language model (LM) frozen, and only training the MLP projector between them.  During this stage, we use simple single-turn dialogues to align the vision token with its corresponding description, meaning, or intention text token. In the second stage, we reduce the batch size to 128, train the LM along with the MLP projector, while keeping the encoder frozen. Regarding data usage, we train the model with a constructed long-thought-chain dataset to help the model develop reasoning abilities for gesture descriptions and abstract meanings. At the same time, we use Mediapipe-extracted landmarks as auxiliary information, added as extra tokens to help the model distinguish fine-grained gestures. The initial learning rates for the two stages are set to 1e-3 and 2e-5, with a warmup ratio of 0.03, and the AdamW optimizer is used with a cosine learning rate schedule. We conduct the training on four 80GB A800 GPUs, with the first stage taking about 7 hours and the second stage taking approximately 14 hours. Throughout both stages, we employ full-parameter
    fine-tuning to ensure maximal model expressivity and performance.

\subsection{Dataset}
\begin{table}[htbp]
  \caption{
    Comparison of gesture recognition datasets. 
    Egocentric/Exocentric columns indicate support for egocentric/exocentric viewpoints.
    Our dataset provides broader coverage with 3 caption types.
  }
  \centering
  \begin{tabularx}{0.95\textwidth}{@{} l *{5}{c} @{}} 
    \toprule
    Dataset & \#Samples & \#Classes & \#Caption Types & Egocentric & Exocentric \\
    \midrule
    Chalearn LAP IsoGD \cite{9172121} & 47,933 & 249 & 1 & $\times$ & $\checkmark$ \\
    HMDB-51 \cite{Kuehne11} & 6,766 & 51 & 1 & $\times$ & $\checkmark$ \\
    Microsoft Kinect \& Leap Motion \cite{7025313} & 1,400 & 10 & 1 & $\times$ & $\checkmark$ \\
    NVIDIA Dynamic Hand Gesture \cite{7780825} & 1,532 & 25 & 1 & $\times$ & $\checkmark$ \\
    EgoGesture  & 24,161 & 83 & 1 & $\checkmark$ & $\times$ \\
    Briareo \cite{manganaro2019hand} & 120 & 12 & 1 & $\times$ & $\checkmark$ \\
    IPN Hand \cite{9412317} & 5,649 & 14 & 1 & $\times$ & $\checkmark$ \\
    JESTER  & 148,092 & 27 & 1 & $\times$ & $\checkmark$ \\
    \textbf{GestureInt (Ours)} & \textbf{159,561} & \textbf{110} & \textbf{3} & $\checkmark$ & $\checkmark$ \\
    \bottomrule
  \end{tabularx}

  \label{tab:dataset_comparison}
\end{table}
We introduce GestureInt, the first dataset specifically designed for gesture meaning and intention understanding. It comprises gesture videos captured from both egocentric (first-person) and exocentric (third-person) viewpoints organized in the form of QA pairs.

GestureInt is build on top of two existing datasets--Jester\cite{9022297} and Egogesture\cite{8299578}, totaling over 150,000 clips across exocentric and egocentric views. GestureInt is constructed via a two-stage annotation pipeline that combines large language model (LLM) generation with expert verification. In Stage 1, all video clips and categories are used to ensure broad coverage. In Stage 2, we focus on 73 intent-bearing categories and construct chain-of-thought (CoT) traces to support intent reasoning.

We structure the dataset in two distinct formats:

1. A three-dimensional separation—where data is explicitly categorized into description, meaning, and intention 

2. A fully integrated format—where these dimensions are logically connected to provide a structured and comprehensive reasoning framework.

The first format enables the model to adapt to the new task  while the second format helps it learn the implicit patterns linking visual features to semantic meanings 

Our dataset includes approximately 130K third-person gesture videos and 30K first-person gesture videos  along with over 400K high-quality textual annotations. A critical flaw in previous gesture datasets is that their annotations are often too brief, lacking the depth necessary for LLMs to accurately learn semantic meanings. Additionally, existing annotations frequently mix descriptions (e.g., “thumbs up”), meanings (e.g., “stop sign”), and intentions (e.g., “make a phone call”)  leading to inconsistencies and ambiguity.

To address these issues, we extended the original labels into detailed depictions and generated possible meanings using ChatGPT \cite{openai2024gpt4technicalreport}  Both steps were reviewed and refined by gesture experts to ensure accuracy and consistency. Furthermore, we expanded the dataset by generating possible gesture intentions related to smart device control and smart interface interactions  making it more applicable to real-world use cases. Several examples of dataset are available in Appendix \ref{app:A.1}.

\begin{figure}
    \centering
    \includegraphics[width=1\linewidth]{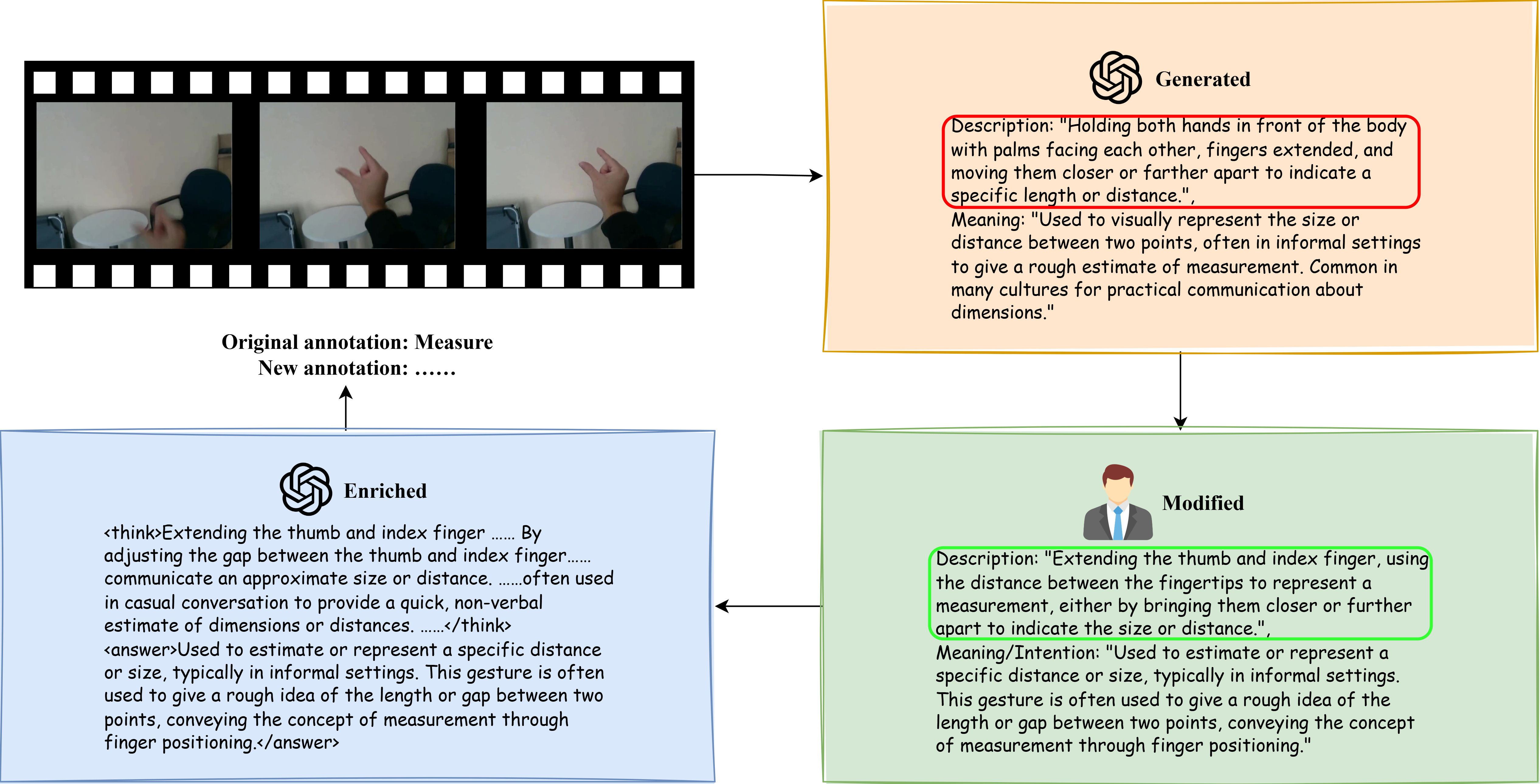}
    \caption{The annotation pipeline}
    \label{fig:enter-label}
\end{figure}
All annotations were produced through a two-stage, LLM-assisted pipeline. In Stage 1, GPT-4o generated a detailed motion description and a plausible semantic meaning for every gesture clip; two domain experts independently verified and amended these outputs and then added the intention label, with disagreements resolved by discussion. In Stage 2, GPT-4o was prompted again to generate a CoT trace that logically connects description, meaning, and intention, after which the same experts checked the trace for logical soundness and refined it when necessary. For completeness, the full prompt used during the data annotation stage is provided in Appendix \ref{app:A.2}.

We split the data as follows: 10 \% of the gesture classes are withheld entirely for open-set testing, while from each of the remaining classes we set aside 10 \% of the videos for closed-set testing and use the rest for training. For instance, the "thumb upward" gesture was held out entirely for open-set evaluation, whereas "thumb downward" videos were included both in training and in closed-set testing.

\subsection{Evaluation Metrics}
To comprehensively evaluate the performance of our gesture intent understanding system, we adopt a set of complementary metrics that measure both the lexical overlap and the semantic consistency between the model predictions and the reference descriptions. Specifically, we report BLEU-1 to BLEU-4, SPICE, and a GPT-assisted semantic accuracy (ACC) as defined below.

\paragraph{BLEU-1 to BLEU-4.}
BLEU (Bilingual Evaluation Understudy) is a precision-based metric commonly used in machine translation and text generation. It measures the n-gram overlap between a candidate sentence and one or more reference sentences. BLEU-n refers to the precision of n-grams, and is formally defined as:
\begin{equation}
\text{BLEU-n} = \text{BP} \cdot \exp\left( \sum_{i=1}^{n} w_i \log p_i \right),
\end{equation}
where $p_i$ is the modified precision for $i$-gram, $w_i$ is typically set to $\frac{1}{n}$, and BP is the brevity penalty to penalize overly short predictions. BLEU-1 measures unigram precision (e.g. BLEU-1 score of 66.65 means 66.65 \% words shared), BLEU-2 tightens the criterion to bigrams such as "thumb up", and so on up to BLEU-4.

\paragraph{SPICE.}
SPICE (Semantic Propositional Image Caption Evaluation) evaluates the semantic content of generated sentences by parsing both candidate and reference sentences into scene graphs consisting of objects, attributes, and relations. It computes an F1 score based on the overlap of these semantic tuples. Unlike BLEU, which relies on surface-level n-gram matching, SPICE is more aligned with human judgments in capturing meaning and conceptual correctness, so paraphrases like "lifts thumb in agreement" still earn high marks.

\paragraph{Semantic Accuracy (ACC).}
To evaluate the semantic correctness of model outputs beyond lexical similarity, we adopt a GPT-4o-assisted evaluation protocol Evaluation and justification of LLM-as-Judge is available in Appendix \ref{app:C}. Specifically, we provide GPT-4o with both the model-generated output and the corresponding ground truth label and prompt it to score the semantic similarity on a 0–5 scale:
\begin{itemize}
    \item 0: completely unrelated
    \item 1-3: weak or partial match
    \item 4–5: semantically accurate
\end{itemize}
We treat a prediction as \textbf{correct} if the score assigned by GPT-4o is greater than or equal to 4. The final ACC score is computed as the percentage of samples classified as correct:
\begin{equation}
\text{ACC} = \frac{N_{\text{score} \geq 4}}{N_{\text{total}}} \times 100\%.
\end{equation}
\subsection{Comparative Experiments}
We evaluate the performance of several state-of-the-art LVLMs on our GestureInt dataset, targeting gesture intention recognition under open-ended, real-world conditions. As shown in Table~\ref{tab:comparison}, our model Gestura consistently surpasses all baselines across both exocentric and egocentric views in terms of intent classification accuracy (ACC) and BLEU scores.

Specifically, Gestura achieves a top-1 intent accuracy of 84.73\% in the exocentric setting and 66.14\% in the egocentric view, substantially outperforming the previous state-of-the-art, GestureGPT, which achieves 72.08\% (closed-set) / 44.46\% (open-set) and 40.07\% (closed-set) / 17.38\% (open-set) under the corresponding settings. This highlights a significant advancement in accurate gesture intent interpretation, particularly under the challenging egocentric perspective.

We further assess each model’s ability to generate descriptive text for gestures using BLEU-1 to BLEU-4 scores. Gestura achieves a BLEU-4 score of 49.83 (exocentric) and 43.67 (egocentric), which are magnitudes higher than those of previous models—e.g., Qwen2.5VL-7B (0.46 and 2.67), or LLaVA-Next-Onevision-7B (0.36 and 0.24). These results highlight Gestura’s superior capacity to translate visual signals into coherent and semantically rich descriptions.

Beyond numeric metrics, Gestura also demonstrates strong qualitative performance. Compared to top-tier models like LLaVA-Next and Qwen2.5VL, which often produce generic or semantically ambiguous outputs, our model generates more context-aware and intention-sensitive descriptions, accurately capturing not just physical movement, but the communicative purpose behind the gesture. This is particularly important in free-form gesture understanding, where similar gestures may convey distinct intents based on subtle variations or situational context.

Overall, our results underscore the effectiveness of combining structured hand landmark information with vision-language pretraining. The substantial gains across both classification and generation tasks validate our two-stage training strategy and landmark-enhanced representation, establishing Gestura as a new state-of-the-art for free-form gesture understanding.
\begin{table}[]
\caption{Comparison between our Gestura and the currently state-of-the-art LVLM models on our Test sets of GestureInt dataset. Metrics include BLEU-1$\sim$4 and ACC (where ACC reports both closed-set / open-set results).}
\footnotesize 
\setlength{\tabcolsep}{3pt} 
\begin{tabular}
{ccccccccccc}
\toprule
\multirow{2}{*}{Method} & \multicolumn{5}{c}{Exocentric}                                        & \multicolumn{5}{c}{Egocentric}                                    \\ \cline{2-11} 
                        & ACC  & BLEU-1         & BLEU-2         & BLEU-3         & BLEU-4     & ACC   & BLEU-1         & BLEU-2         & BLEU-3         & BLEU-4       \\ \hline
Internvl2.5-8B  \cite{chen2025expandingperformanceboundariesopensource}      & 8.29   & 12.07          & 3.95           & 1.12           & 0.34      &   10.55    & 14.18          & 3.76           & 1.03           & 0.25        \\
LLaVA-Next-Video-7B-DPO  \cite{liu2024llavanext}  & 8.20   & 13.01          & 4.72           & 1.20           & 0.34     &  8.41     & 13.59          & 4.76           & 1.26           & 0.36       \\
LLaVA-Next-Onevision-7B  \cite{li2024llavaonevisioneasyvisualtask}&10.97    & 16.84          & 3.86           & 1.04           & 0.36      & 12.28     & 13.40          & 3.66           & 1.00           & 0.24         \\
LLaVA-Video-7B      \cite{zhang2024videoinstructiontuningsynthetic}   &  11.77   & 18.13          & 4.18           & 1.14           & 0.38     & 13.90     & 20.43          & 5.24           & 1.58           & 0.51         \\
Qwen2.5VL-7B        \cite{bai2025qwen25vltechnicalreport}   &  6.60     & 17.33           & 5.85           & 1.57            & 0.46     & 5.42      & 19.41           & 9.53           & 4.80           & 2.67        \\
Video-LLaVA-7B       \cite{lin2024videollavalearningunitedvisual}    &2.89      & 12.50          & 1.95           & 0.24           & 0.04     &6.50      & 13.93          & 2.35           & 0.33           & 0.07      \\
GestureGPT(close)*        & 72.08      & 23.53          & 19.10           & 16.63           & 15.05     &40.07      & 21.26          & 15.14           & 12.33           & 10.75      \\
GestureGPT(open)*        & 44.46      & 18.65          & 12.21           & 9.11           & 6.79     &17.38      & 21.11          & 11.74           & 6.95           & 4.39      \\
\textbf{Gestura(close)}    & \textbf{84.73}    & \textbf{53.94} & \textbf{51.87} & \textbf{50.61} & \textbf{49.83} & \textbf{66.14} & \textbf{52.33} & \textbf{47.72} & \textbf{45.15} & \textbf{43.67} \\ 
\textbf{Gestura(open)}    & \textbf{65.65}    & \textbf{34.88} & \textbf{25.36} & \textbf{19.37 } & \textbf{14.17} & \textbf{21.71} & \textbf{33.73} & \textbf{22.10} & \textbf{14.60} & \textbf{9.93} \\ 
\bottomrule
\end{tabular}

\label{tab:comparison}
\end{table}


\begin{table}[H]
\caption{Ablation analysis of Qwen2.5VL and Gestura under \textbf{exocentric} settings.}
\begin{tabular}{cccccccc}
\toprule
\multirow{2}{*}{Method} & \multicolumn{6}{c}{Exocentric}                        \\ \cline{3-8} 
\multicolumn{2}{c}{}               & ACC          & BLEU-1 & BLEU-2 & BLEU-3 & BLEU-4 & SPICE   \\ \hline
\multirow{2}{*}{Qwen2.5VL-7B (Same data)}  & close& 73.46  & 45.16  & 42.13  & 40.58  & 39.68  & 0.50  \\
                                    & open  & 66.76 & 29.82  & 21.19  & 16.16  & 11.57  & 0.26  \\ \hline
\multirow{2}{*}{\textbf{Gestura}}           & close  & 84.73  & 53.94  & 51.87  & 50.61  & 49.83  & 0.63 \\
                                    & open & 65.65  & 34.88  & 25.36  & 19.37  & 14.17  & 0.30 \\ \bottomrule
\end{tabular}

\label{tab:exo_ablation}
\end{table}

\begin{table}[H]
\caption{Ablation analysis of Qwen2.5VL and Gestura under \textbf{egocentric} settings.}
\begin{tabular}{cccccccc}
\toprule
\multirow{2}{*}{Method} & \multicolumn{6}{c}{Egocentric}                    \\ \cline{3-8} 
\multicolumn{2}{c}{}            & ACC              & BLEU-1 & BLEU-2 & BLEU-3 & BLEU-4 & SPICE  \\ \hline
\multirow{2}{*}{Qwen2.5VL-7B (Same data)}  & close  & 41.39& 39.11  & 31.41  & 27.16  & 25.06  & 0.36  \\
                                    & open  & 21.23  & 31.41  & 18.55  & 10.81  & 5.93   & 0.17 \\ \hline
\multirow{2}{*}{\textbf{Gestura}}           & close  & 66.14 & 52.33  & 47.72  & 45.15  & 43.67  & 0.56 \\
                                    & open & 21.71  & 33.73  & 22.10  & 14.60  & 9.93   & 0.20  \\ \bottomrule
\end{tabular}

\label{tab:ego_ablation}
\end{table}

\subsection{Ablation Study}
To better understand the contributions of each component in our system, we conduct a comprehensive ablation study under both exocentric and egocentric test settings. Specifically, we evaluate three configurations: (1) baseline, (2) fine-tuning with only Chain-of-Thought (CoT), (3) full pipeline with both CoT and LPM used during both training and inference, and (4) fine-tuning with both CoT and Landmark Processing Module (LPM) but testing without it. The performance is compared in both closed-set and open-set scenarios.

\paragraph{Ablation Study Results.}
Table~\ref{tab:ablation} presents a comprehensive ablation study evaluating the contributions of Chain-of-Thought (CoT) and the Landmark Processing Module (LPM) under both exocentric and egocentric settings, across closed-set and open-set scenarios.

We observe that incorporating both CoT and LPM achieves the best overall performance. Specifically, the full model configuration (row 3) reaches the highest intent recognition accuracy in both views: 84.73\% (exocentric closed-set) and 66.14\% (egocentric closed-set). In comparison, the CoT-only setup (row 2) yields lower accuracy (80.27\% and 60.34\%, respectively), while the baseline without any of these components performs significantly worse (70.13\% and 53.14\%). These results demonstrate that LPM offers complementary structural cues that enhance the semantic grounding facilitated by CoT reasoning.

In the open-set setting, where gesture categories are unseen during training, overall performance drops due to the increased challenge of generalization. Nevertheless, our model remains robust: the full configuration still outperforms all alternatives, achieving 65.65\% (exocentric) and 21.71\% (egocentric). 

Considering that the extraction and encoding of Landmark information in practical applications may negatively affect latency, we compare the performance during the test phase with and without using MediaPipe information. Notably, even when LPM is removed at test time (row 4), the model maintains competitive accuracy, highlighting that gesture-aware fine-tuning with LPM imparts lasting benefits even when the module is no longer available during inference. Our experiments demonstrate two key findings: (1) Training with landmark information improves model robustness; (2) The model retains high performance even when landmark data is excluded during testing.

Additionally, to assess the effectiveness of our proposed framework, we conducted an ablation study by fine-tuning the latest multimodal model Qwen2.5VL-7B on the same data used by Gestura. As shown in Tables~\ref{tab:exo_ablation} and~\ref{tab:ego_ablation}, despite being trained on identical datasets, Gestura significantly outperforms Qwen2.5VL across both exocentric and egocentric settings. In both closed and open scenarios, Gestura achieves higher accuracy, BLEU scores, and SPICE metrics, demonstrating its superior capability in understanding free-form gestures and inferring intent.

These findings confirm that:
\begin{enumerate}
    \item CoT reasoning can better connect gesture descriptions with their meanings or intentions by providing an objective and logical inference process. This approach enables the model to make reasonable predictions for novel gestures by leveraging existing prior knowledge.
\item Gesture landmarks offer both direct (test-time input) and indirect (training-time regularization) advantages, improving robustness and generalizability. 
\item The combination of CoT and LPM provides the most reliable and interpretable results across diverse perspectives and settings, underscoring the importance of integrating both symbolic and spatial priors for free-form gesture understanding.
\item Compared to state-of-the-art LVLM fine-tuned on the same dataset, our framework consistently achieves superior performance across exocentric and egocentric views. This highlights the effectiveness of our framework and the advantage of a structured, intent-aware pipeline.
\end{enumerate}

\begin{table*}[!]
\centering
\caption{Ablation Study on ACC results under four ablation experimental conditions across both \textbf{Exocentric} and \textbf{Egocentric} scenarios.}
\label{tab:ablation}
\begin{tabular}{ccccc|cccc}
\toprule
\multirow{2}{*}{Methods} & \multicolumn{4}{c}{Components} &\multicolumn{2}{c}{Exocentric} & \multicolumn{2}{c}{Egocentric}  \\
\cline{2-9} &\textit{Multi-View.}
&\textit{CoT.} & \textit{LPM fine-tuning.}  &\textit{LPM testing.}~&\textit{close}&\textit{open}&\textit{close}&\textit{open} \\
\hline
1 & - & - & - & -   & 2.89  & / & 6.50  & /  \\
2 & \checkmark & - & - & -   & 70.13  & 25.34&53.14  & 12.25  \\
3 & \checkmark & \checkmark & - & -  & 80.27  & 63.52 & 60.34 & 17.78 \\
4 & \checkmark & \checkmark & \checkmark & \checkmark  & \textbf{84.73}  &65.65 & 66.14  & \textbf{21.71}  \\
\textbf{Gestura} & \checkmark & \checkmark & \checkmark & -  & 82.73  &  \textbf{66.48} & \textbf{66.20}  & 21.06  \\

\bottomrule
\end{tabular}
\end{table*}

\section{Implementation}

In this section, we explore the deployment of Gestura in a device-server hybrid architecture, where the model is hosted on a backend server and communicates with a wearable AI glasses prototype. This setup balances computational efficiency and real-time responsiveness, while maintaining low latency and user mobility.

We describe our system implementation, including the hardware setup of the AI glasses and the interaction pipeline from gesture capture to audio feedback. To evaluate real-world usability, we conducted a user study in which participants performed intuitive, free-form gestures based on given intent prompts. We further assess the system’s performance through metrics such as intent recognition accuracy and response time, demonstrating that Gestura achieves reliable on-the-fly inference. Finally, feature-level analysis reveals how semantic reasoning helps unify diverse gesture expressions, reinforcing the importance of integrating visual and language understanding in gesture-intent recognition.

\subsection{System Setup}
The implementation of Gestura for on-device deployment is mainly on our prototype AI glasses. We implement the model first, and using the AI glasses as the wearable device to handle the visual input with its monocular camera in the middle of the frame, and the audio input with its microphone. After receiving, the inputs go directly to the model for processing, the output text go straight to TTS tool and the transformed audio sent to the AI glasses and played by the embedded speaker. Specifically, Gestura was deployed on a NVIDIA A100 GPU for inferencing. The AI glasses equipped Qualcomm HexagonTM NPU, with a lightweight Linux-based operating system run on it. Video input was captured via an embedded camera with a resolution of 4K resolution at 30 FPS, ensuring compatibility with real-world gesture capture scenarios.\\
\textbf{Task Description}: Specific eight tasks assigned to participants related to smart device and interface control. Participants are required to execute gesture freely based on provided key words about tasks.\\
\textbf{Task Procedure}: Participants were first given a brief introduction to the smart device and interface control task without any prior guidance on specific gestures. They then wore our device and proceeded with the experiment. Each participant was provided with eight independent keyword phrases and was instructed to spontaneously perform the most natural and intuitive gesture corresponding to each phrase, without prior contemplation. The system recorded the experimental data based on the feedback received from the device.

\subsection{Data Collection and Participants}
To evaluate the on-device deployment, we conducted an additional user study involving 13 volunteers. We recruited 13 adult volunteers (9 men, 4 women) from an institute of a company.
Their ages ranged from 20 to 30 years (MEAN = 23.8, SD = 4.2). Gesture-interaction experience was assessed with a
pre-study questionnaire, which shows that seven participants have prior gesture-based interface experience while others
don’t.

Participants were presented with a set of 8 intent-related gesture keywords and were instructed to perform free-form gestures based on their intuitive understanding of each keyword, without receiving any predefined demonstrations or constraints. This setup aimed to capture naturalistic and diverse gesture expressions across individuals.

In total, the study yielded 104 gesture samples (13 participants × 8 keywords), each recorded as a first-person perspective video with an average duration of approximately 2 seconds. All recordings were collected under consistent lighting and environmental conditions to ensure data quality and comparability. This participant-generated dataset serves as a valuable complement to our open-source gesture-intent corpus, offering realistic and unconstrained gesture expressions for evaluating real-time recognition performance in wearable scenarios.

\subsection{Evaluation}

The performance of the on-device \textit{Gestura} system was evaluated by incorporating a language-based intent inference approach. Specifically, we utilized ChatGPT as the intent recognition intermediary: for each gesture video, our model first generated a textual output, which was subsequently fed into ChatGPT to infer the most likely corresponding function.

For each sample, we collected the Top-1, Top-3, and Top-5 intent predictions based on ChatGPT's semantic ranking against a predefined intent pool. Recognition accuracy was then computed as the percentage of cases in which the ground-truth intent appeared within the top-$k$ predictions ($k \in \{1, 3, 5\}$).

In addition to intent recognition accuracy, we assessed the system using the following metrics:

\begin{itemize}
    \item \textbf{Top-$k$ Accuracy}: The proportion of test samples where the correct intent appeared in the top-1, top-3, or top-5 predicted intents.
    \item \textbf{Response Time}: The average latency (in milliseconds) from video input to final intent prediction, capturing the total runtime of the vision-to-language-to-intent pipeline.
\end{itemize}

These evaluation metrics reflect a comprehensive balance between recognition accuracy, responsiveness, and deployment efficiency, addressing the practical considerations of real-world edge computing scenarios.

\subsection{Overall Performance}

\begin{table}[htbp] 
  \caption{Top-1 / Top-3 / Top-5 accuracy per intent category in an open-world experiment.} 
  \centering 
  \begin{tabular}{cccccc}
    \toprule
    \textbf{Intent Category} & \textbf{Samples} & \textbf{Top-1 Acc} & \textbf{Top-3 Acc} & \textbf{Top-5 Acc} \\
    \hline
    come over / beckon & 13 & 0.0000  & 0.3846 & 0.7692 \\
    confirm / agree & 13 & 0.7692  & 1.0000 & 1.0000 \\
    decrease temp (AC) & 13 & 0.1538 & 0.3846 & 0.3846 \\
    go to next page & 13 & 0.3077  & 0.5385 & 0.6154 \\
    go to previous page & 13 & 0.4615  & 0.7692 & 0.8462 \\
    increase temp (AC) & 13 & 0.0000  & 0.0769 & 0.1538 \\
    make a phone call & 13 & 0.6154  & 0.7692 & 0.9231 \\
    take a photo & 13 & 0.6923  & 0.6923 & 0.8462 \\
    \hline
    \textbf{Overall} & \textbf{104} & \textbf{0.3750} & \textbf{0.5769} & \textbf{0.6923} \\
    \bottomrule 
  \end{tabular}
  \label{tab:gesture_accuracy} 
\end{table}

Based on the results presented in Table~\ref{tab:gesture_accuracy}, our system demonstrates promising performance on egocentric gesture-intent recognition, achieving an overall Top-1 accuracy of 37.5\%, Top-3 accuracy of 57.7\%, and Top-5 accuracy of 69.2\% across all eight intent categories. These results highlight the system’s effectiveness in capturing intent-relevant semantics, even when users perform free-form and visually diverse gestures. In terms of latency, the system achieves a practical response time ranging from 5 to 10 seconds, with an average of \textbf{7.83 seconds}. Notably, when excluding communication delays and TTS overhead, the core model processes inputs and produces outputs within an average of just \textbf{1.6 seconds}—a substantial improvement over the multi-agent-based GestureGPT, which requires 227 seconds per gesture. This dramatic reduction in response time is largely attributed to our lightweight, end-to-end framework design, which eliminates the need for complex inter-agent communication and enables much faster and more efficient inference.

Intent categories such as “confirm / agree” and “make a phone call” achieve particularly high recognition rates (Top-1 accuracy of 76.9\% and 61.5\% respectively), suggesting that these gestures tend to be more consistent across users. In contrast, categories like “increase temp (AC)” and “come over / beckon” exhibit lower Top-1 performance, reflecting greater variability in individual expression and ambiguity in visual cues. However, the Top-3 and Top-5 accuracies indicate that the correct intent is still frequently ranked among the model’s top predictions, which is valuable for downstream interactive applications.

\begin{figure}[H]
    \centering
    \includegraphics[width=0.5\linewidth]{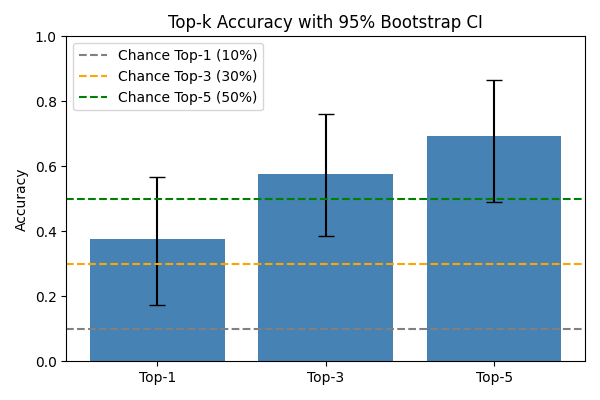}
    \caption{Top-k Accuracy with 95\% Bootstrap CI}
    \label{fig:topk_errorbar}
\end{figure}

We provide an error-bar plot \ref{fig:topk_errorbar} contrasting our Top-1/3/5 accuracies with their chance baselines. One-sample t-tests show that Top-1(\textit{p} = 0.0386 < 0.05, \textit{d} = 0.90) and Top-3 (p =  \textit{0.0309, \textit{d} = 0.95} outperform chance significantly, whereas Top-5 accuracy showed a positive trend (\textit{p} = 0.1036 $\approx$ 0.10).

To strengthen the statistical interpretation, we additionally performed bootstrap analysis with 10,000 resamples, yielding consistent results.
For Top-5, the bootstrap delta was still positive (+19.2 pp), and the effect size remained moderate (
\textit{d}
= 0.66), the 95\% CI marginally included zero [–1.0 pp, +36.5 pp]. This suggests a positive trend despite the lack of statistical significance, reinforcing the overall robustness of the model under looser evaluation thresholds.
The Top-1 improvement over chance was +27.5 pp (95\% BCa CI = [7.3 pp, 46.7 pp], 
\textit{d} = 0.90), and Top-3 was +27.7 pp (CI = [8.5 pp, 46.0 pp], 
\textit{d} = 0.95). These results confirm that the model is statistically and practically effective in producing correct Top-1 and Top-3 predictions.

Overall, these results demonstrate the robustness and adaptability of our approach in realistic, unconstrained egocentric scenarios, and underscore the benefits of integrating visual and semantic understanding for gesture-based intent inference.


\subsection{Analysis}
The variability of real-world gesture performance can introduce a gap between controlled experimental data and practical applications. Categories with semantic uniqueness tend to achieve better performance.

Misdirections in the open-world test arise mainly from three factors. (1) Cultural and personal variation: volunteers often select different gestures to express the same intent. (2) Motion symmetry: visually similar trajectories such as “hand moving up” versus “hand moving down” remain hard to disambiguate. This situation appears as well when we generating enriched motion description using GPT-4o (3) Gesture repetition: users frequently repeat dynamic motions for emphasis, which obscures directionality and confuses opposite pairs like “scroll up vs. scroll down” or “swipe left vs. swipe right.” 

\begin{figure}[H]
  \centering
  \begin{minipage}[t]{0.45\textwidth}
    \centering
    \includegraphics[width=\linewidth]{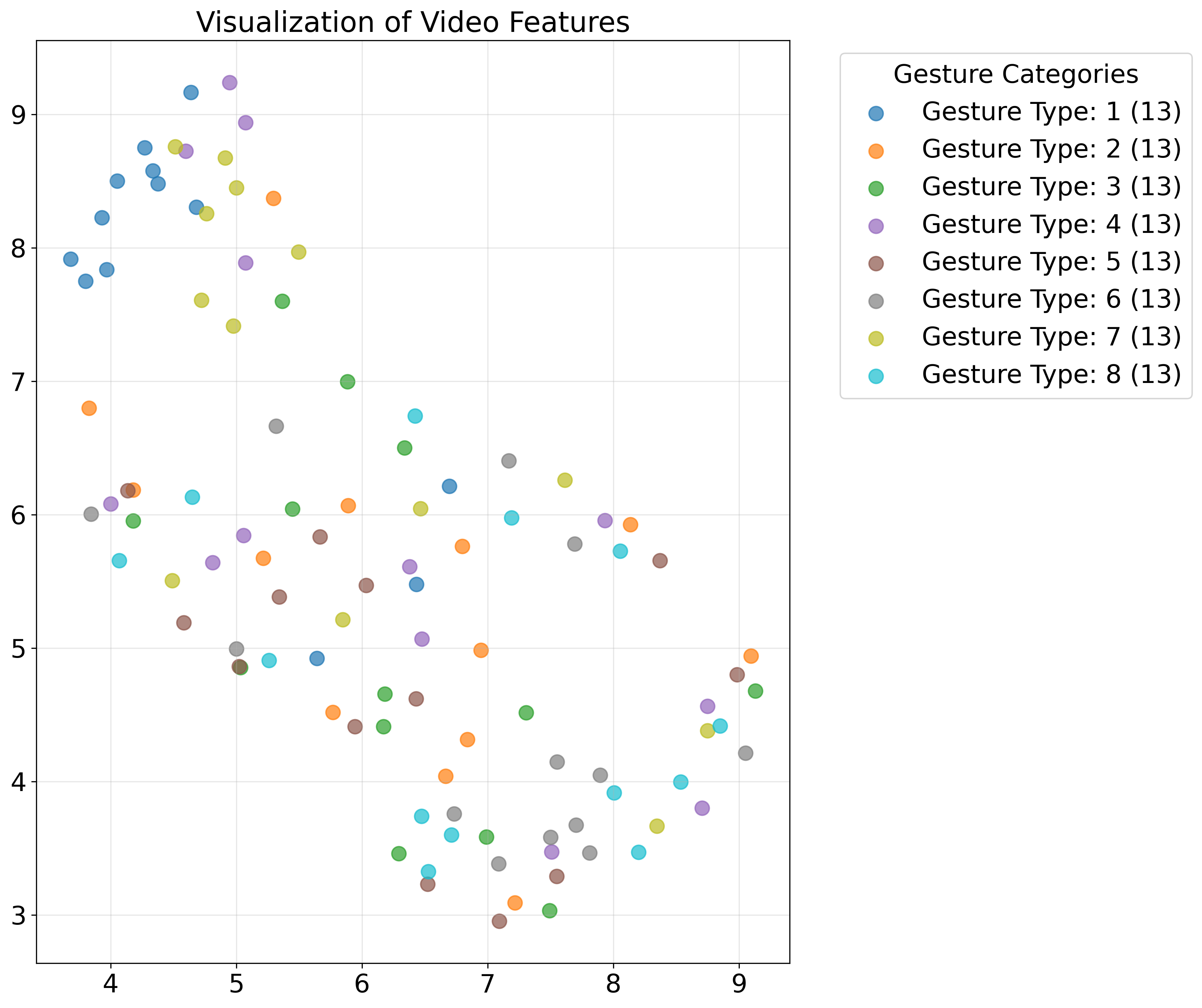}
    \caption{Visualization of Video Features}
    \label{fig:video_features}
  \end{minipage}
  \hfill
  \begin{minipage}[t]{0.45\textwidth}
    \centering
    \includegraphics[width=\linewidth]{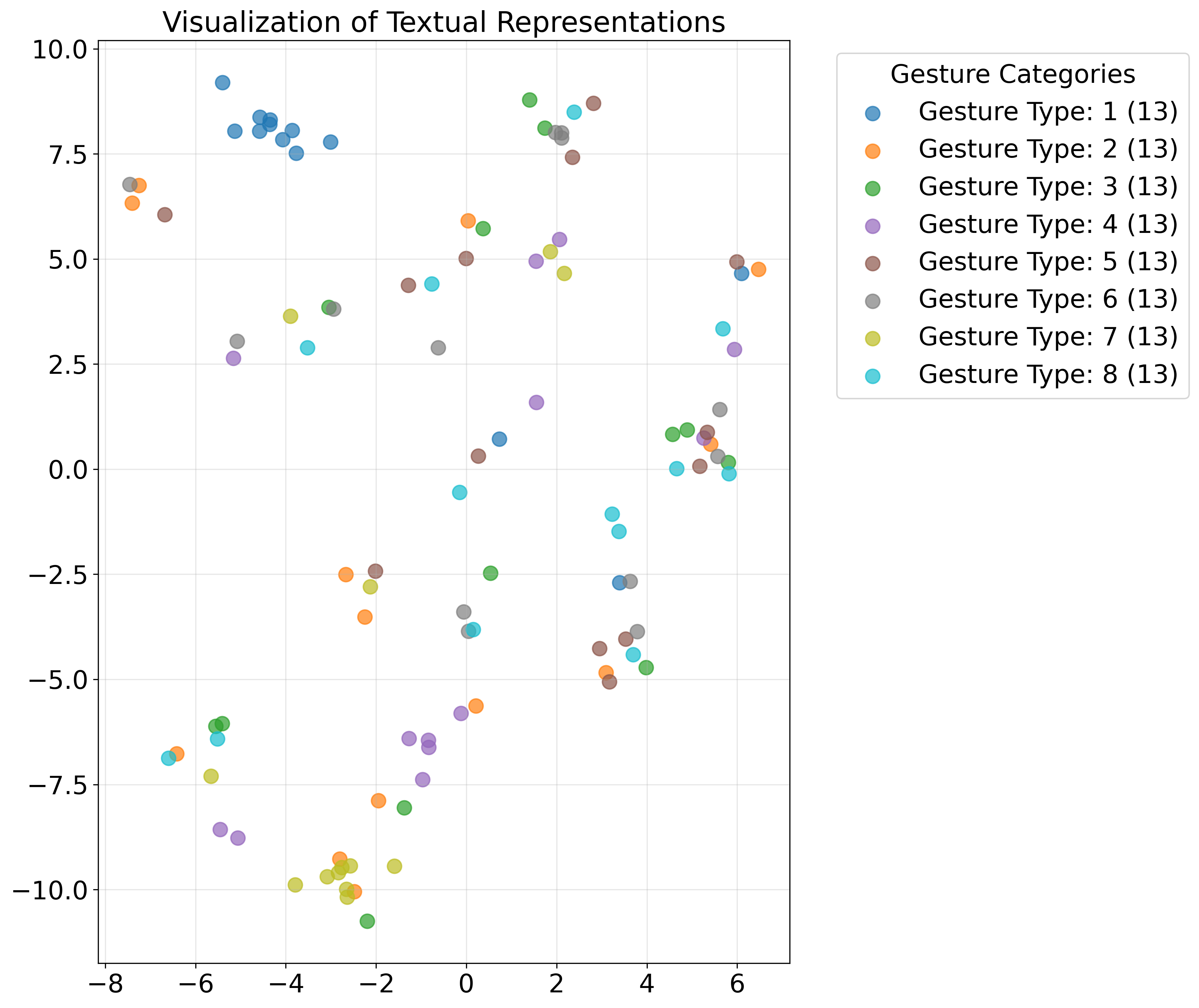}
    \caption{Visualization of Textual Representations}
    \label{fig:text_features}
  \end{minipage}
\end{figure}

In real-world experiments, we observed that participants often performed different gestures to express the same intent. As illustrated in Figure~\ref{fig:video_features}, the visual features extracted by the video encoder — represented as tensors of shape (8, 4096) — tend to exhibit clustering only when the performed gestures are visually similar, such as those corresponding to Gesture Type 1 and Gesture Type 7. However, for other gesture types, due to individual variations in motor expression, the extracted features are more dispersed, leading to less distinct clusters.

In contrast, as shown in Figure~\ref{fig:text_features}, the semantic features derived from LLM output tokens — typically in the shape of (56, 4096) —  demonstrate clearer clustering patterns even when the visual gestures differ. By leveraging our proposed pipeline, gestures that differ significantly in visual appearance yet share the same underlying intent are effectively brought closer together in the semantic embedding space. These findings highlight the importance of gesture-intent recognition, as it enables more robust understanding beyond surface-level gesture differences.
\section{Discussion}

Despite the progress made in free-form gesture comprehension, a persistent challenge remains: how can open-set gesture recognition systems approach the performance of closed-set counterparts? In closed-set settings, models operate under the assumption that all gesture categories are predefined and known during training, leading to high accuracy but limited generalization. Conversely, open-set systems must generalize to unseen or user-defined gestures, which inherently introduces ambiguity and performance degradation.

The performance gap between exocentric closed-set and egocentric open-set scenarios arises from several factors. First, the exocentric subset contains fewer categories but more training samples per class, making it easier for the model to generalize. In contrast, the egocentric set exhibits greater category diversity and dynamic hand motions, increasing ambiguity. Second, in open-world settings, the model may force predictions into seen categories even when visual matches are poor—particularly problematic for egocentric gestures with more varied semantics. Third, egocentric videos are often affected by environmental noise such as lighting or cluttered backgrounds. Lastly, human performance variability (e.g., due to cultural factors or directional ambiguity) further compounds the recognition challenge.

While Gestura is currently deployed on a single NVIDIA A100 40GB GPU for benchmarking, its architecture is designed with efficiency in mind. Given its 8B parameter scale and streamlined end-to-end design, there is strong potential for edge deployment. Notably, with optimization techniques such as model quantization, pruning, and distillation, Gestura can be significantly compressed without substantial performance loss. For instance, platforms like the NVIDIA Jetson AGX Orin\cite{nvidiajetson}, which provides up to 64GB of shared memory and 2048-core Ampere GPU, offer a feasible hardware base for deploying a compressed version of our model. This opens the door for real-time, free-form gesture understanding on edge devices, enabling applications in smart homes, AR/VR systems, and robotics without reliance on cloud resources.

To bridge this gap, we propose several future directions:

\textbf{1) Enriching high-quality, semantically-annotated gesture data:} Our current framework benefits from a structured, QA-formatted gesture-intent dataset. Expanding this dataset with more diverse users, environments, and cultural contexts would significantly enhance model robustness. Incorporating multimodal annotations—e.g., combining gesture videos with eye gaze, speech cues, or physiological signals—could provide richer context for disambiguation.

\textbf{2) Adaptive inference through test-time scaling:} Instead of relying solely on fixed training representations, test-time adaptation techniques (e.g., retrieval-augmented inference, few-shot adaptation, or gradient-based prompt tuning) may allow models to dynamically adjust to novel gestures. These approaches enable the system to scale semantically at inference time without retraining, better capturing the latent intent behind unfamiliar hand motions.

\textbf{3) Contrastive grounding with weak supervision:} Leveraging large-scale, weakly labeled video data (e.g., instructional videos, sign language corpora) via contrastive pretraining may help align free-form gestures with high-level semantics in the absence of dense annotation. This could reduce the reliance on curated datasets while enhancing open-set generalization.

While Gestura is currently deployed on a single NVIDIA A100 40GB GPU for benchmarking, its architecture is designed with efficiency and scalability in mind. Given its 8B parameter scale and streamlined end-to-end design, there is strong potential for edge deployment. Notably, with optimization techniques such as model quantization, pruning, and knowledge distillation, Gestura can be significantly compressed without substantial performance loss. For instance, platforms like the NVIDIA Jetson AGX Orin, which offers up to 64GB of shared memory and a 2048-core Ampere GPU, provide a viable hardware foundation for running a lightweight version of the model. This opens the door for real-time, free-form gesture understanding on edge devices, enabling applications in smart homes, AR/VR systems, and robotics without reliance on cloud resources.

Ultimately, we envision a future where open-set gesture recognition systems not only narrow the performance gap with their closed-set counterparts but also excel in adaptability, efficiency, and deployability. Realizing this vision requires synergistic advancements in robust gesture encoding, semantic grounding, and context-aware reasoning, alongside hardware-aware model design. With frameworks like Gestura, which balance performance and efficiency, we are one step closer to enabling real-time, intuitive, and accessible gesture interaction across diverse real-world environments.

\section{Conclusion}

In this work, we introduce Gestura, a free-form hand gesture understanding system that harnesses the power of LVLMs to bridge the gap between motion patterns and semantic intent. Our framework is built upon a novel hierarchical architecture that combines landmark processing with advanced cross-modal alignment techniques. This enables the system to not only interpret subtle hand movements but also to infer the underlying user intent with remarkable accuracy. Through extensive experiments in real-world scenarios, Gestura demonstrates exceptional performance, achieving a top-1 intent accuracy of 84.73\% in exocentric settings and 66.14\% in egocentric settings. These results highlight the system's ability to handle the complexity and variability of open-world environments. Moreover, Gestura's average response time of just 7.83 seconds, with pure model inference taking only 1.6 seconds, underscores its suitability for real-time applications. By eliminating the need for users to learn predefined gestures, Gestura sets a new standard for intuitive and natural human-computer interaction, opening avenues for future research in creating more adaptive and user-friendly interfaces.
\section{Acknowledgments}
We thank the AI Glasses product team at China Telecom for their enthusiastic support and valuable assistance.








\bibliographystyle{ACM-Reference-Format}
\bibliography{sample-base}


\begin{thebibliography}{47}


\ifx \showCODEN    \undefined \def \showCODEN     #1{\unskip}     \fi
\ifx \showISBNx    \undefined \def \showISBNx     #1{\unskip}     \fi
\ifx \showISBNxiii \undefined \def \showISBNxiii  #1{\unskip}     \fi
\ifx \showISSN     \undefined \def \showISSN      #1{\unskip}     \fi
\ifx \showLCCN     \undefined \def \showLCCN      #1{\unskip}     \fi
\ifx \shownote     \undefined \def \shownote      #1{#1}          \fi
\ifx \showarticletitle \undefined \def \showarticletitle #1{#1}   \fi
\ifx \showURL      \undefined \def \showURL       {\relax}        \fi
\providecommand\bibfield[2]{#2}
\providecommand\bibinfo[2]{#2}
\providecommand\natexlab[1]{#1}
\providecommand\showeprint[2][]{arXiv:#2}

\bibitem[Aich et~al\mbox{.}(2023)]%
        {aichDataFreeClassIncrementalHand2023}
\bibfield{author}{\bibinfo{person}{Shubhra Aich}, \bibinfo{person}{Jesus {Ruiz-Santaquiteria}}, \bibinfo{person}{Zhenyu Lu}, \bibinfo{person}{Prachi Garg}, \bibinfo{person}{K~J Joseph}, \bibinfo{person}{Alvaro~Fernandez Garcia}, \bibinfo{person}{Vineeth~N Balasubramanian}, \bibinfo{person}{Kenrick Kin}, \bibinfo{person}{Chengde Wan}, \bibinfo{person}{Necati~Cihan Camgoz}, \bibinfo{person}{Shugao Ma}, {and} \bibinfo{person}{Fernando De~La~Torre}.} \bibinfo{year}{2023}\natexlab{}.
\newblock \showarticletitle{Data-{{Free Class-Incremental Hand Gesture Recognition}}}. In \bibinfo{booktitle}{\emph{2023 {{IEEE}}/{{CVF International Conference}} on {{Computer Vision}} ({{ICCV}})}}. \bibinfo{publisher}{IEEE}, \bibinfo{address}{Paris, France}, \bibinfo{pages}{20901--20910}.
\newblock
\showISBNx{979-8-3503-0718-4}
\href{https://doi.org/10.1109/ICCV51070.2023.01916}{doi:\nolinkurl{10.1109/ICCV51070.2023.01916}}


\bibitem[Al~Mudawi et~al\mbox{.}(2024)]%
        {almudawiInnovativeHealthcareSolutions2024}
\bibfield{author}{\bibinfo{person}{Naif Al~Mudawi}, \bibinfo{person}{Hira Ansar}, \bibinfo{person}{Abdulwahab Alazeb}, \bibinfo{person}{Hanan Aljuaid}, \bibinfo{person}{Yahay AlQahtani}, \bibinfo{person}{Asaad Algarni}, \bibinfo{person}{Ahmad Jalal}, {and} \bibinfo{person}{Hui Liu}.} \bibinfo{year}{2024}\natexlab{}.
\newblock \showarticletitle{Innovative Healthcare Solutions: Robust Hand Gesture Recognition of Daily Life Routines Using {{1D CNN}}}.
\newblock \bibinfo{journal}{\emph{Frontiers in Bioengineering and Biotechnology}}  \bibinfo{volume}{12} (\bibinfo{date}{July} \bibinfo{year}{2024}), \bibinfo{pages}{1401803}.
\newblock
\showISSN{2296-4185}
\href{https://doi.org/10.3389/fbioe.2024.1401803}{doi:\nolinkurl{10.3389/fbioe.2024.1401803}}


\bibitem[Ao et~al\mbox{.}(2023)]%
        {aoGestureDiffuCLIPGestureDiffusion2023}
\bibfield{author}{\bibinfo{person}{Tenglong Ao}, \bibinfo{person}{Zeyi Zhang}, {and} \bibinfo{person}{Libin Liu}.} \bibinfo{year}{2023}\natexlab{}.
\newblock \showarticletitle{{{GestureDiffuCLIP}}: {{Gesture Diffusion Model}} with {{CLIP Latents}}}.
\newblock \bibinfo{journal}{\emph{ACM Transactions on Graphics}} \bibinfo{volume}{42}, \bibinfo{number}{4} (\bibinfo{date}{Aug.} \bibinfo{year}{2023}), \bibinfo{pages}{1--18}.
\newblock
\showISSN{0730-0301, 1557-7368}
\showLCCN{A{\textbar}6.2{\textbar}Q1{\textbar}1区TOP}
\href{https://doi.org/10.1145/3592097}{doi:\nolinkurl{10.1145/3592097}}


\bibitem[Bai et~al\mbox{.}(2025)]%
        {bai2025qwen25vltechnicalreport}
\bibfield{author}{\bibinfo{person}{Shuai Bai}, \bibinfo{person}{Keqin Chen}, \bibinfo{person}{Xuejing Liu}, \bibinfo{person}{Jialin Wang}, \bibinfo{person}{Wenbin Ge}, \bibinfo{person}{Sibo Song}, \bibinfo{person}{Kai Dang}, \bibinfo{person}{Peng Wang}, \bibinfo{person}{Shijie Wang}, \bibinfo{person}{Jun Tang}, \bibinfo{person}{Humen Zhong}, \bibinfo{person}{Yuanzhi Zhu}, \bibinfo{person}{Mingkun Yang}, \bibinfo{person}{Zhaohai Li}, \bibinfo{person}{Jianqiang Wan}, \bibinfo{person}{Pengfei Wang}, \bibinfo{person}{Wei Ding}, \bibinfo{person}{Zheren Fu}, \bibinfo{person}{Yiheng Xu}, \bibinfo{person}{Jiabo Ye}, \bibinfo{person}{Xi Zhang}, \bibinfo{person}{Tianbao Xie}, \bibinfo{person}{Zesen Cheng}, \bibinfo{person}{Hang Zhang}, \bibinfo{person}{Zhibo Yang}, \bibinfo{person}{Haiyang Xu}, {and} \bibinfo{person}{Junyang Lin}.} \bibinfo{year}{2025}\natexlab{}.
\newblock \bibinfo{title}{Qwen2.5-VL Technical Report}.
\newblock
\showeprint[arxiv]{2502.13923}~[cs.CV]
\urldef\tempurl%
\url{https://arxiv.org/abs/2502.13923}
\showURL{%
\tempurl}


\bibitem[Bandini and Zariffa(2022)]%
        {bandiniAnalysisHandsEgocentric2022}
\bibfield{author}{\bibinfo{person}{Andrea Bandini} {and} \bibinfo{person}{Jos{\'e} Zariffa}.} \bibinfo{year}{2022}\natexlab{}.
\newblock \bibinfo{title}{Analysis of the Hands in Egocentric Vision: {{A}} Survey}.
\newblock
\showLCCN{arXiv}
\href{https://doi.org/10.48550/arXiv.1912.10867}{doi:\nolinkurl{10.48550/arXiv.1912.10867}}
\showeprint[arxiv]{1912.10867}


\bibitem[Benitez-Garcia et~al\mbox{.}(2021)]%
        {9412317}
\bibfield{author}{\bibinfo{person}{Gibran Benitez-Garcia}, \bibinfo{person}{Jesus Olivares-Mercado}, \bibinfo{person}{Gabriel Sanchez-Perez}, {and} \bibinfo{person}{Keiji Yanai}.} \bibinfo{year}{2021}\natexlab{}.
\newblock \showarticletitle{{ IPN Hand: A Video Dataset and Benchmark for Real-Time Continuous Hand Gesture Recognition }}. In \bibinfo{booktitle}{\emph{2020 25th International Conference on Pattern Recognition (ICPR)}}. \bibinfo{publisher}{IEEE Computer Society}, \bibinfo{address}{Los Alamitos, CA, USA}, \bibinfo{pages}{4340--4347}.
\newblock
\showISSN{1051-4651}
\href{https://doi.org/10.1109/ICPR48806.2021.9412317}{doi:\nolinkurl{10.1109/ICPR48806.2021.9412317}}


\bibitem[Chang et~al\mbox{.}(2021)]%
        {10.1145/3463509}
\bibfield{author}{\bibinfo{person}{Yuhu Chang}, \bibinfo{person}{Yingying Zhao}, \bibinfo{person}{Mingzhi Dong}, \bibinfo{person}{Yujiang Wang}, \bibinfo{person}{Yutian Lu}, \bibinfo{person}{Qin Lv}, \bibinfo{person}{Robert~P. Dick}, \bibinfo{person}{Tun Lu}, \bibinfo{person}{Ning Gu}, {and} \bibinfo{person}{Li Shang}.} \bibinfo{year}{2021}\natexlab{}.
\newblock \showarticletitle{MemX: An Attention-Aware Smart Eyewear System for Personalized Moment Auto-capture}.
\newblock \bibinfo{journal}{\emph{Proc. ACM Interact. Mob. Wearable Ubiquitous Technol.}} \bibinfo{volume}{5}, \bibinfo{number}{2}, Article \bibinfo{articleno}{56} (\bibinfo{date}{June} \bibinfo{year}{2021}), \bibinfo{numpages}{23}~pages.
\newblock
\href{https://doi.org/10.1145/3463509}{doi:\nolinkurl{10.1145/3463509}}


\bibitem[Chen et~al\mbox{.}(2025)]%
        {chen2025expandingperformanceboundariesopensource}
\bibfield{author}{\bibinfo{person}{Zhe Chen}, \bibinfo{person}{Weiyun Wang}, \bibinfo{person}{Yue Cao}, \bibinfo{person}{Yangzhou Liu}, \bibinfo{person}{Zhangwei Gao}, \bibinfo{person}{Erfei Cui}, \bibinfo{person}{Jinguo Zhu}, \bibinfo{person}{Shenglong Ye}, \bibinfo{person}{Hao Tian}, \bibinfo{person}{Zhaoyang Liu}, \bibinfo{person}{Lixin Gu}, \bibinfo{person}{Xuehui Wang}, \bibinfo{person}{Qingyun Li}, \bibinfo{person}{Yimin Ren}, \bibinfo{person}{Zixuan Chen}, \bibinfo{person}{Jiapeng Luo}, \bibinfo{person}{Jiahao Wang}, \bibinfo{person}{Tan Jiang}, \bibinfo{person}{Bo Wang}, \bibinfo{person}{Conghui He}, \bibinfo{person}{Botian Shi}, \bibinfo{person}{Xingcheng Zhang}, \bibinfo{person}{Han Lv}, \bibinfo{person}{Yi Wang}, \bibinfo{person}{Wenqi Shao}, \bibinfo{person}{Pei Chu}, \bibinfo{person}{Zhongying Tu}, \bibinfo{person}{Tong He}, \bibinfo{person}{Zhiyong Wu}, \bibinfo{person}{Huipeng Deng}, \bibinfo{person}{Jiaye Ge}, \bibinfo{person}{Kai Chen}, \bibinfo{person}{Kaipeng Zhang}, \bibinfo{person}{Limin
  Wang}, \bibinfo{person}{Min Dou}, \bibinfo{person}{Lewei Lu}, \bibinfo{person}{Xizhou Zhu}, \bibinfo{person}{Tong Lu}, \bibinfo{person}{Dahua Lin}, \bibinfo{person}{Yu Qiao}, \bibinfo{person}{Jifeng Dai}, {and} \bibinfo{person}{Wenhai Wang}.} \bibinfo{year}{2025}\natexlab{}.
\newblock \bibinfo{title}{Expanding Performance Boundaries of Open-Source Multimodal Models with Model, Data, and Test-Time Scaling}.
\newblock
\showeprint[arxiv]{2412.05271}~[cs.CV]
\urldef\tempurl%
\url{https://arxiv.org/abs/2412.05271}
\showURL{%
\tempurl}


\bibitem[DeepSeek-AI et~al\mbox{.}(2025)]%
        {deepseekai2025deepseekr1incentivizingreasoningcapability}
\bibfield{author}{\bibinfo{person}{DeepSeek-AI}, \bibinfo{person}{Daya Guo}, \bibinfo{person}{Dejian Yang}, \bibinfo{person}{Haowei Zhang}, \bibinfo{person}{Junxiao Song}, \bibinfo{person}{Ruoyu Zhang}, \bibinfo{person}{Runxin Xu}, \bibinfo{person}{Qihao Zhu}, \bibinfo{person}{Shirong Ma}, \bibinfo{person}{Peiyi Wang}, \bibinfo{person}{Xiao Bi}, \bibinfo{person}{Xiaokang Zhang}, \bibinfo{person}{Xingkai Yu}, \bibinfo{person}{Yu Wu}, \bibinfo{person}{Z.~F. Wu}, \bibinfo{person}{Zhibin Gou}, \bibinfo{person}{Zhihong Shao}, \bibinfo{person}{Zhuoshu Li}, \bibinfo{person}{Ziyi Gao}, \bibinfo{person}{Aixin Liu}, \bibinfo{person}{Bing Xue}, \bibinfo{person}{Bingxuan Wang}, \bibinfo{person}{Bochao Wu}, \bibinfo{person}{Bei Feng}, \bibinfo{person}{Chengda Lu}, \bibinfo{person}{Chenggang Zhao}, \bibinfo{person}{Chengqi Deng}, \bibinfo{person}{Chenyu Zhang}, \bibinfo{person}{Chong Ruan}, \bibinfo{person}{Damai Dai}, \bibinfo{person}{Deli Chen}, \bibinfo{person}{Dongjie Ji}, \bibinfo{person}{Erhang Li},
  \bibinfo{person}{Fangyun Lin}, \bibinfo{person}{Fucong Dai}, \bibinfo{person}{Fuli Luo}, \bibinfo{person}{Guangbo Hao}, \bibinfo{person}{Guanting Chen}, \bibinfo{person}{Guowei Li}, \bibinfo{person}{H. Zhang}, \bibinfo{person}{Han Bao}, \bibinfo{person}{Hanwei Xu}, \bibinfo{person}{Haocheng Wang}, \bibinfo{person}{Honghui Ding}, \bibinfo{person}{Huajian Xin}, \bibinfo{person}{Huazuo Gao}, \bibinfo{person}{Hui Qu}, \bibinfo{person}{Hui Li}, \bibinfo{person}{Jianzhong Guo}, \bibinfo{person}{Jiashi Li}, \bibinfo{person}{Jiawei Wang}, \bibinfo{person}{Jingchang Chen}, \bibinfo{person}{Jingyang Yuan}, \bibinfo{person}{Junjie Qiu}, \bibinfo{person}{Junlong Li}, \bibinfo{person}{J.~L. Cai}, \bibinfo{person}{Jiaqi Ni}, \bibinfo{person}{Jian Liang}, \bibinfo{person}{Jin Chen}, \bibinfo{person}{Kai Dong}, \bibinfo{person}{Kai Hu}, \bibinfo{person}{Kaige Gao}, \bibinfo{person}{Kang Guan}, \bibinfo{person}{Kexin Huang}, \bibinfo{person}{Kuai Yu}, \bibinfo{person}{Lean Wang}, \bibinfo{person}{Lecong Zhang},
  \bibinfo{person}{Liang Zhao}, \bibinfo{person}{Litong Wang}, \bibinfo{person}{Liyue Zhang}, \bibinfo{person}{Lei Xu}, \bibinfo{person}{Leyi Xia}, \bibinfo{person}{Mingchuan Zhang}, \bibinfo{person}{Minghua Zhang}, \bibinfo{person}{Minghui Tang}, \bibinfo{person}{Meng Li}, \bibinfo{person}{Miaojun Wang}, \bibinfo{person}{Mingming Li}, \bibinfo{person}{Ning Tian}, \bibinfo{person}{Panpan Huang}, \bibinfo{person}{Peng Zhang}, \bibinfo{person}{Qiancheng Wang}, \bibinfo{person}{Qinyu Chen}, \bibinfo{person}{Qiushi Du}, \bibinfo{person}{Ruiqi Ge}, \bibinfo{person}{Ruisong Zhang}, \bibinfo{person}{Ruizhe Pan}, \bibinfo{person}{Runji Wang}, \bibinfo{person}{R.~J. Chen}, \bibinfo{person}{R.~L. Jin}, \bibinfo{person}{Ruyi Chen}, \bibinfo{person}{Shanghao Lu}, \bibinfo{person}{Shangyan Zhou}, \bibinfo{person}{Shanhuang Chen}, \bibinfo{person}{Shengfeng Ye}, \bibinfo{person}{Shiyu Wang}, \bibinfo{person}{Shuiping Yu}, \bibinfo{person}{Shunfeng Zhou}, \bibinfo{person}{Shuting Pan}, \bibinfo{person}{S.~S. Li},
  \bibinfo{person}{Shuang Zhou}, \bibinfo{person}{Shaoqing Wu}, \bibinfo{person}{Shengfeng Ye}, \bibinfo{person}{Tao Yun}, \bibinfo{person}{Tian Pei}, \bibinfo{person}{Tianyu Sun}, \bibinfo{person}{T. Wang}, \bibinfo{person}{Wangding Zeng}, \bibinfo{person}{Wanjia Zhao}, \bibinfo{person}{Wen Liu}, \bibinfo{person}{Wenfeng Liang}, \bibinfo{person}{Wenjun Gao}, \bibinfo{person}{Wenqin Yu}, \bibinfo{person}{Wentao Zhang}, \bibinfo{person}{W.~L. Xiao}, \bibinfo{person}{Wei An}, \bibinfo{person}{Xiaodong Liu}, \bibinfo{person}{Xiaohan Wang}, \bibinfo{person}{Xiaokang Chen}, \bibinfo{person}{Xiaotao Nie}, \bibinfo{person}{Xin Cheng}, \bibinfo{person}{Xin Liu}, \bibinfo{person}{Xin Xie}, \bibinfo{person}{Xingchao Liu}, \bibinfo{person}{Xinyu Yang}, \bibinfo{person}{Xinyuan Li}, \bibinfo{person}{Xuecheng Su}, \bibinfo{person}{Xuheng Lin}, \bibinfo{person}{X.~Q. Li}, \bibinfo{person}{Xiangyue Jin}, \bibinfo{person}{Xiaojin Shen}, \bibinfo{person}{Xiaosha Chen}, \bibinfo{person}{Xiaowen Sun}, \bibinfo{person}{Xiaoxiang
  Wang}, \bibinfo{person}{Xinnan Song}, \bibinfo{person}{Xinyi Zhou}, \bibinfo{person}{Xianzu Wang}, \bibinfo{person}{Xinxia Shan}, \bibinfo{person}{Y.~K. Li}, \bibinfo{person}{Y.~Q. Wang}, \bibinfo{person}{Y.~X. Wei}, \bibinfo{person}{Yang Zhang}, \bibinfo{person}{Yanhong Xu}, \bibinfo{person}{Yao Li}, \bibinfo{person}{Yao Zhao}, \bibinfo{person}{Yaofeng Sun}, \bibinfo{person}{Yaohui Wang}, \bibinfo{person}{Yi Yu}, \bibinfo{person}{Yichao Zhang}, \bibinfo{person}{Yifan Shi}, \bibinfo{person}{Yiliang Xiong}, \bibinfo{person}{Ying He}, \bibinfo{person}{Yishi Piao}, \bibinfo{person}{Yisong Wang}, \bibinfo{person}{Yixuan Tan}, \bibinfo{person}{Yiyang Ma}, \bibinfo{person}{Yiyuan Liu}, \bibinfo{person}{Yongqiang Guo}, \bibinfo{person}{Yuan Ou}, \bibinfo{person}{Yuduan Wang}, \bibinfo{person}{Yue Gong}, \bibinfo{person}{Yuheng Zou}, \bibinfo{person}{Yujia He}, \bibinfo{person}{Yunfan Xiong}, \bibinfo{person}{Yuxiang Luo}, \bibinfo{person}{Yuxiang You}, \bibinfo{person}{Yuxuan Liu}, \bibinfo{person}{Yuyang Zhou},
  \bibinfo{person}{Y.~X. Zhu}, \bibinfo{person}{Yanhong Xu}, \bibinfo{person}{Yanping Huang}, \bibinfo{person}{Yaohui Li}, \bibinfo{person}{Yi Zheng}, \bibinfo{person}{Yuchen Zhu}, \bibinfo{person}{Yunxian Ma}, \bibinfo{person}{Ying Tang}, \bibinfo{person}{Yukun Zha}, \bibinfo{person}{Yuting Yan}, \bibinfo{person}{Z.~Z. Ren}, \bibinfo{person}{Zehui Ren}, \bibinfo{person}{Zhangli Sha}, \bibinfo{person}{Zhe Fu}, \bibinfo{person}{Zhean Xu}, \bibinfo{person}{Zhenda Xie}, \bibinfo{person}{Zhengyan Zhang}, \bibinfo{person}{Zhewen Hao}, \bibinfo{person}{Zhicheng Ma}, \bibinfo{person}{Zhigang Yan}, \bibinfo{person}{Zhiyu Wu}, \bibinfo{person}{Zihui Gu}, \bibinfo{person}{Zijia Zhu}, \bibinfo{person}{Zijun Liu}, \bibinfo{person}{Zilin Li}, \bibinfo{person}{Ziwei Xie}, \bibinfo{person}{Ziyang Song}, \bibinfo{person}{Zizheng Pan}, \bibinfo{person}{Zhen Huang}, \bibinfo{person}{Zhipeng Xu}, \bibinfo{person}{Zhongyu Zhang}, {and} \bibinfo{person}{Zhen Zhang}.} \bibinfo{year}{2025}\natexlab{}.
\newblock \bibinfo{title}{DeepSeek-R1: Incentivizing Reasoning Capability in LLMs via Reinforcement Learning}.
\newblock
\showeprint[arxiv]{2501.12948}~[cs.CL]
\urldef\tempurl%
\url{https://arxiv.org/abs/2501.12948}
\showURL{%
\tempurl}


\bibitem[Dosovitskiy et~al\mbox{.}(2020)]%
        {DBLP:journals/corr/abs-2010-11929}
\bibfield{author}{\bibinfo{person}{Alexey Dosovitskiy}, \bibinfo{person}{Lucas Beyer}, \bibinfo{person}{Alexander Kolesnikov}, \bibinfo{person}{Dirk Weissenborn}, \bibinfo{person}{Xiaohua Zhai}, \bibinfo{person}{Thomas Unterthiner}, \bibinfo{person}{Mostafa Dehghani}, \bibinfo{person}{Matthias Minderer}, \bibinfo{person}{Georg Heigold}, \bibinfo{person}{Sylvain Gelly}, \bibinfo{person}{Jakob Uszkoreit}, {and} \bibinfo{person}{Neil Houlsby}.} \bibinfo{year}{2020}\natexlab{}.
\newblock \showarticletitle{An Image is Worth 16x16 Words: Transformers for Image Recognition at Scale}.
\newblock \bibinfo{journal}{\emph{CoRR}}  \bibinfo{volume}{abs/2010.11929} (\bibinfo{year}{2020}).
\newblock
\showeprint[arXiv]{2010.11929}
\urldef\tempurl%
\url{https://arxiv.org/abs/2010.11929}
\showURL{%
\tempurl}


\bibitem[Feng et~al\mbox{.}(2023)]%
        {10.5555/3666122.3669222}
\bibfield{author}{\bibinfo{person}{Guhao Feng}, \bibinfo{person}{Bohang Zhang}, \bibinfo{person}{Yuntian Gu}, \bibinfo{person}{Haotian Ye}, \bibinfo{person}{Di He}, {and} \bibinfo{person}{Liwei Wang}.} \bibinfo{year}{2023}\natexlab{}.
\newblock \showarticletitle{Towards revealing the mystery behind chain of thought: a theoretical perspective}. In \bibinfo{booktitle}{\emph{Proceedings of the 37th International Conference on Neural Information Processing Systems}} (New Orleans, LA, USA) \emph{(\bibinfo{series}{NIPS '23})}. \bibinfo{publisher}{Curran Associates Inc.}, \bibinfo{address}{Red Hook, NY, USA}, Article \bibinfo{articleno}{3100}, \bibinfo{numpages}{42}~pages.
\newblock


\bibitem[Higger et~al\mbox{.}({[n.\,d.]})]%
        {higgerOpenWorldHumanRobotInteraction}
\bibfield{author}{\bibinfo{person}{Mark Higger}, \bibinfo{person}{Polina Rygina}, \bibinfo{person}{Logan Daigler}, \bibinfo{person}{Lara~Ferreira Bezerra}, \bibinfo{person}{Zhao Han}, {and} \bibinfo{person}{Tom Williams}.} \bibinfo{year}{[n.\,d.]}\natexlab{}.
\newblock \showarticletitle{Toward {{Open-World Human-Robot Interaction}}: {{What Types}} of {{Gestures Are Used}} in {{Task-Based Open-World Referential Communication}}?}
\newblock  (\bibinfo{year}{[n.\,d.]}).
\newblock
\showLCCN{Invalid Title}


\bibitem[Hu et~al\mbox{.}(2024)]%
        {huIOTeethIntraOralTeeth2024}
\bibfield{author}{\bibinfo{person}{Zhizhang Hu}, \bibinfo{person}{Amirmohammad Radmehr}, \bibinfo{person}{Yue Zhang}, \bibinfo{person}{Shijia Pan}, {and} \bibinfo{person}{Phuc Nguyen}.} \bibinfo{year}{2024}\natexlab{}.
\newblock \showarticletitle{{{IOTeeth}}: {{Intra-Oral Teeth Sensing System}} for {{Dental Occlusal Diseases Recognition}}}.
\newblock \bibinfo{journal}{\emph{Proceedings of the ACM on Interactive, Mobile, Wearable and Ubiquitous Technologies}} \bibinfo{volume}{8}, \bibinfo{number}{1} (\bibinfo{date}{March} \bibinfo{year}{2024}), \bibinfo{pages}{1--29}.
\newblock
\showISSN{2474-9567}
\href{https://doi.org/10.1145/3643516}{doi:\nolinkurl{10.1145/3643516}}


\bibitem[Huang et~al\mbox{.}(2024)]%
        {huangSpeciFingersFingerIdentification2024}
\bibfield{author}{\bibinfo{person}{Zeyuan Huang}, \bibinfo{person}{Cangjun Gao}, \bibinfo{person}{Haiyan Wang}, \bibinfo{person}{Xiaoming Deng}, \bibinfo{person}{Yu-Kun Lai}, \bibinfo{person}{Cuixia Ma}, \bibinfo{person}{Sheng-feng Qin}, \bibinfo{person}{Yong-Jin Liu}, {and} \bibinfo{person}{Hongan Wang}.} \bibinfo{year}{2024}\natexlab{}.
\newblock \showarticletitle{{{SpeciFingers}}: {{Finger Identification}} and {{Error Correction}} on {{Capacitive Touchscreens}}}.
\newblock \bibinfo{journal}{\emph{Proceedings of the ACM on Interactive, Mobile, Wearable and Ubiquitous Technologies}} \bibinfo{volume}{8}, \bibinfo{number}{1} (\bibinfo{date}{March} \bibinfo{year}{2024}), \bibinfo{pages}{1--28}.
\newblock
\showISSN{2474-9567}
\href{https://doi.org/10.1145/3643559}{doi:\nolinkurl{10.1145/3643559}}


\bibitem[Khaokaew et~al\mbox{.}(2024)]%
        {khaokaewMAPLEMobileApp2024}
\bibfield{author}{\bibinfo{person}{Yonchanok Khaokaew}, \bibinfo{person}{Hao Xue}, {and} \bibinfo{person}{Flora~D. Salim}.} \bibinfo{year}{2024}\natexlab{}.
\newblock \showarticletitle{{{MAPLE}}: {{Mobile App Prediction Leveraging Large Language Model Embeddings}}}.
\newblock \bibinfo{journal}{\emph{Proceedings of the ACM on Interactive, Mobile, Wearable and Ubiquitous Technologies}} \bibinfo{volume}{8}, \bibinfo{number}{1} (\bibinfo{date}{March} \bibinfo{year}{2024}), \bibinfo{pages}{1--25}.
\newblock
\showISSN{2474-9567}
\href{https://doi.org/10.1145/3643514}{doi:\nolinkurl{10.1145/3643514}}


\bibitem[King et~al\mbox{.}(2024)]%
        {kingSashaCreativeGoalOriented2024}
\bibfield{author}{\bibinfo{person}{Evan King}, \bibinfo{person}{Haoxiang Yu}, \bibinfo{person}{Sangsu Lee}, {and} \bibinfo{person}{Christine Julien}.} \bibinfo{year}{2024}\natexlab{}.
\newblock \showarticletitle{Sasha: {{Creative Goal-Oriented Reasoning}} in {{Smart Homes}} with {{Large Language Models}}}.
\newblock \bibinfo{journal}{\emph{Proceedings of the ACM on Interactive, Mobile, Wearable and Ubiquitous Technologies}} \bibinfo{volume}{8}, \bibinfo{number}{1} (\bibinfo{date}{March} \bibinfo{year}{2024}), \bibinfo{pages}{1--38}.
\newblock
\showISSN{2474-9567}
\href{https://doi.org/10.1145/3643505}{doi:\nolinkurl{10.1145/3643505}}


\bibitem[Kuehne et~al\mbox{.}(2011)]%
        {Kuehne11}
\bibfield{author}{\bibinfo{person}{H. Kuehne}, \bibinfo{person}{H. Jhuang}, \bibinfo{person}{E. Garrote}, \bibinfo{person}{T. Poggio}, {and} \bibinfo{person}{T. Serre}.} \bibinfo{year}{2011}\natexlab{}.
\newblock \showarticletitle{{HMDB}: a large video database for human motion recognition}. In \bibinfo{booktitle}{\emph{Proceedings of the International Conference on Computer Vision (ICCV)}}.
\newblock


\bibitem[Li et~al\mbox{.}(2024)]%
        {li2024llavaonevisioneasyvisualtask}
\bibfield{author}{\bibinfo{person}{Bo Li}, \bibinfo{person}{Yuanhan Zhang}, \bibinfo{person}{Dong Guo}, \bibinfo{person}{Renrui Zhang}, \bibinfo{person}{Feng Li}, \bibinfo{person}{Hao Zhang}, \bibinfo{person}{Kaichen Zhang}, \bibinfo{person}{Peiyuan Zhang}, \bibinfo{person}{Yanwei Li}, \bibinfo{person}{Ziwei Liu}, {and} \bibinfo{person}{Chunyuan Li}.} \bibinfo{year}{2024}\natexlab{}.
\newblock \bibinfo{title}{LLaVA-OneVision: Easy Visual Task Transfer}.
\newblock
\showeprint[arxiv]{2408.03326}~[cs.CV]
\urldef\tempurl%
\url{https://arxiv.org/abs/2408.03326}
\showURL{%
\tempurl}


\bibitem[Lin et~al\mbox{.}(2024)]%
        {lin2024videollavalearningunitedvisual}
\bibfield{author}{\bibinfo{person}{Bin Lin}, \bibinfo{person}{Yang Ye}, \bibinfo{person}{Bin Zhu}, \bibinfo{person}{Jiaxi Cui}, \bibinfo{person}{Munan Ning}, \bibinfo{person}{Peng Jin}, {and} \bibinfo{person}{Li Yuan}.} \bibinfo{year}{2024}\natexlab{}.
\newblock \bibinfo{title}{Video-LLaVA: Learning United Visual Representation by Alignment Before Projection}.
\newblock
\showeprint[arxiv]{2311.10122}~[cs.CV]
\urldef\tempurl%
\url{https://arxiv.org/abs/2311.10122}
\showURL{%
\tempurl}


\bibitem[Linardakis et~al\mbox{.}(2025)]%
        {linardakisSurveyHandGesture2025}
\bibfield{author}{\bibinfo{person}{Manousos Linardakis}, \bibinfo{person}{Iraklis Varlamis}, {and} \bibinfo{person}{Georgios~Th Papadopoulos}.} \bibinfo{year}{2025}\natexlab{}.
\newblock \bibinfo{title}{Survey on {{Hand Gesture Recognition}} from {{Visual Input}}}.
\newblock
\showLCCN{arXiv}
\href{https://doi.org/10.48550/arXiv.2501.11992}{doi:\nolinkurl{10.48550/arXiv.2501.11992}}
\showeprint[arxiv]{2501.11992}~[cs]


\bibitem[Liu et~al\mbox{.}(2024a)]%
        {liuTagSleep3DRFbased3D2024}
\bibfield{author}{\bibinfo{person}{Chen Liu}, \bibinfo{person}{Zixuan Dong}, \bibinfo{person}{Li Huang}, \bibinfo{person}{Wenlong Yan}, \bibinfo{person}{Xin Wang}, \bibinfo{person}{Dingyi Fang}, {and} \bibinfo{person}{Xiaojiang Chen}.} \bibinfo{year}{2024}\natexlab{a}.
\newblock \showarticletitle{{{TagSleep3D}}: {{RF-based 3D Sleep Posture Skeleton Recognition}}}.
\newblock \bibinfo{journal}{\emph{Proceedings of the ACM on Interactive, Mobile, Wearable and Ubiquitous Technologies}} \bibinfo{volume}{8}, \bibinfo{number}{1} (\bibinfo{date}{March} \bibinfo{year}{2024}), \bibinfo{pages}{1--28}.
\newblock
\showISSN{2474-9567}
\href{https://doi.org/10.1145/3643512}{doi:\nolinkurl{10.1145/3643512}}


\bibitem[Liu et~al\mbox{.}(2024b)]%
        {liu2024llavanext}
\bibfield{author}{\bibinfo{person}{Haotian Liu}, \bibinfo{person}{Chunyuan Li}, \bibinfo{person}{Yuheng Li}, \bibinfo{person}{Bo Li}, \bibinfo{person}{Yuanhan Zhang}, \bibinfo{person}{Sheng Shen}, {and} \bibinfo{person}{Yong~Jae Lee}.} \bibinfo{year}{2024}\natexlab{b}.
\newblock \bibinfo{title}{LLaVA-NeXT: Improved reasoning, OCR, and world knowledge}.
\newblock
\urldef\tempurl%
\url{https://llava-vl.github.io/blog/2024-01-30-llava-next/}
\showURL{%
\tempurl}


\bibitem[Liu et~al\mbox{.}(2023)]%
        {10.5555/3666122.3667638}
\bibfield{author}{\bibinfo{person}{Haotian Liu}, \bibinfo{person}{Chunyuan Li}, \bibinfo{person}{Qingyang Wu}, {and} \bibinfo{person}{Yong~Jae Lee}.} \bibinfo{year}{2023}\natexlab{}.
\newblock \showarticletitle{Visual instruction tuning}. In \bibinfo{booktitle}{\emph{Proceedings of the 37th International Conference on Neural Information Processing Systems}} (New Orleans, LA, USA) \emph{(\bibinfo{series}{NIPS '23})}. \bibinfo{publisher}{Curran Associates Inc.}, \bibinfo{address}{Red Hook, NY, USA}, Article \bibinfo{articleno}{1516}, \bibinfo{numpages}{25}~pages.
\newblock


\bibitem[Lugaresi et~al\mbox{.}(2019)]%
        {lugaresi2019mediapipeframeworkbuildingperception}
\bibfield{author}{\bibinfo{person}{Camillo Lugaresi}, \bibinfo{person}{Jiuqiang Tang}, \bibinfo{person}{Hadon Nash}, \bibinfo{person}{Chris McClanahan}, \bibinfo{person}{Esha Uboweja}, \bibinfo{person}{Michael Hays}, \bibinfo{person}{Fan Zhang}, \bibinfo{person}{Chuo-Ling Chang}, \bibinfo{person}{Ming~Guang Yong}, \bibinfo{person}{Juhyun Lee}, \bibinfo{person}{Wan-Teh Chang}, \bibinfo{person}{Wei Hua}, \bibinfo{person}{Manfred Georg}, {and} \bibinfo{person}{Matthias Grundmann}.} \bibinfo{year}{2019}\natexlab{}.
\newblock \bibinfo{title}{MediaPipe: A Framework for Building Perception Pipelines}.
\newblock
\showeprint[arxiv]{1906.08172}~[cs.DC]
\urldef\tempurl%
\url{https://arxiv.org/abs/1906.08172}
\showURL{%
\tempurl}


\bibitem[Manganaro et~al\mbox{.}(2019)]%
        {manganaro2019hand}
\bibfield{author}{\bibinfo{person}{Fabio Manganaro}, \bibinfo{person}{Stefano Pini}, \bibinfo{person}{Guido Borghi}, \bibinfo{person}{Roberto Vezzani}, {and} \bibinfo{person}{Rita Cucchiara}.} \bibinfo{year}{2019}\natexlab{}.
\newblock \showarticletitle{Hand Gestures for the Human-Car Interaction: the Briareo dataset}. In \bibinfo{booktitle}{\emph{International Conference on Image Analysis and Processing}}. Springer, \bibinfo{pages}{560--571}.
\newblock


\bibitem[Marin et~al\mbox{.}(2014)]%
        {7025313}
\bibfield{author}{\bibinfo{person}{Giulio Marin}, \bibinfo{person}{Fabio Dominio}, {and} \bibinfo{person}{Pietro Zanuttigh}.} \bibinfo{year}{2014}\natexlab{}.
\newblock \showarticletitle{Hand gesture recognition with leap motion and kinect devices}. In \bibinfo{booktitle}{\emph{2014 IEEE International Conference on Image Processing (ICIP)}}. \bibinfo{pages}{1565--1569}.
\newblock
\href{https://doi.org/10.1109/ICIP.2014.7025313}{doi:\nolinkurl{10.1109/ICIP.2014.7025313}}


\bibitem[Materzynska et~al\mbox{.}(2019)]%
        {9022297}
\bibfield{author}{\bibinfo{person}{Joanna Materzynska}, \bibinfo{person}{Guillaume Berger}, \bibinfo{person}{Ingo Bax}, {and} \bibinfo{person}{Roland Memisevic}.} \bibinfo{year}{2019}\natexlab{}.
\newblock \showarticletitle{The Jester Dataset: A Large-Scale Video Dataset of Human Gestures}. In \bibinfo{booktitle}{\emph{2019 IEEE/CVF International Conference on Computer Vision Workshop (ICCVW)}}. \bibinfo{pages}{2874--2882}.
\newblock
\href{https://doi.org/10.1109/ICCVW.2019.00349}{doi:\nolinkurl{10.1109/ICCVW.2019.00349}}


\bibitem[Miah et~al\mbox{.}(2023)]%
        {miahDynamicHandGesture2023}
\bibfield{author}{\bibinfo{person}{Abu Saleh~Musa Miah}, \bibinfo{person}{Md. Al~Mehedi Hasan}, {and} \bibinfo{person}{Jungpil Shin}.} \bibinfo{year}{2023}\natexlab{}.
\newblock \showarticletitle{Dynamic {{Hand Gesture Recognition Using Multi-Branch Attention Based Graph}} and {{General Deep Learning Model}}}.
\newblock \bibinfo{journal}{\emph{IEEE Access}}  \bibinfo{volume}{11} (\bibinfo{year}{2023}), \bibinfo{pages}{4703--4716}.
\newblock
\showISSN{2169-3536}
\href{https://doi.org/10.1109/ACCESS.2023.3235368}{doi:\nolinkurl{10.1109/ACCESS.2023.3235368}}


\bibitem[Molchanov et~al\mbox{.}(2016)]%
        {7780825}
\bibfield{author}{\bibinfo{person}{Pavlo Molchanov}, \bibinfo{person}{Xiaodong Yang}, \bibinfo{person}{Shalini Gupta}, \bibinfo{person}{Kihwan Kim}, \bibinfo{person}{Stephen Tyree}, {and} \bibinfo{person}{Jan Kautz}.} \bibinfo{year}{2016}\natexlab{}.
\newblock \showarticletitle{Online Detection and Classification of Dynamic Hand Gestures with Recurrent 3D Convolutional Neural Networks}. In \bibinfo{booktitle}{\emph{2016 IEEE Conference on Computer Vision and Pattern Recognition (CVPR)}}. \bibinfo{pages}{4207--4215}.
\newblock
\href{https://doi.org/10.1109/CVPR.2016.456}{doi:\nolinkurl{10.1109/CVPR.2016.456}}


\bibitem[NVIDIA({[n.\,d.]})]%
        {nvidiajetson}
\bibfield{author}{\bibinfo{person}{NVIDIA}.} \bibinfo{year}{[n.\,d.]}\natexlab{}.
\newblock \bibinfo{title}{NVIDIA Jetson Orin Next-level AI performance for next-gen robotics and edge solutions.}
\newblock
\urldef\tempurl%
\url{https://www.nvidia.com/en-us/autonomous-machines/embedded-systems/jetson-orin/}
\showURL{%
\tempurl}


\bibitem[OpenAI et~al\mbox{.}(2024)]%
        {openai2024gpt4technicalreport}
\bibfield{author}{\bibinfo{person}{OpenAI}, \bibinfo{person}{Josh Achiam}, \bibinfo{person}{Steven Adler}, \bibinfo{person}{Sandhini Agarwal}, \bibinfo{person}{Lama Ahmad}, \bibinfo{person}{Ilge Akkaya}, \bibinfo{person}{Florencia~Leoni Aleman}, \bibinfo{person}{Diogo Almeida}, \bibinfo{person}{Janko Altenschmidt}, \bibinfo{person}{Sam Altman}, \bibinfo{person}{Shyamal Anadkat}, \bibinfo{person}{Red Avila}, \bibinfo{person}{Igor Babuschkin}, \bibinfo{person}{Suchir Balaji}, \bibinfo{person}{Valerie Balcom}, \bibinfo{person}{Paul Baltescu}, \bibinfo{person}{Haiming Bao}, \bibinfo{person}{Mohammad Bavarian}, \bibinfo{person}{Jeff Belgum}, \bibinfo{person}{Irwan Bello}, \bibinfo{person}{Jake Berdine}, \bibinfo{person}{Gabriel Bernadett-Shapiro}, \bibinfo{person}{Christopher Berner}, \bibinfo{person}{Lenny Bogdonoff}, \bibinfo{person}{Oleg Boiko}, \bibinfo{person}{Madelaine Boyd}, \bibinfo{person}{Anna-Luisa Brakman}, \bibinfo{person}{Greg Brockman}, \bibinfo{person}{Tim Brooks}, \bibinfo{person}{Miles Brundage},
  \bibinfo{person}{Kevin Button}, \bibinfo{person}{Trevor Cai}, \bibinfo{person}{Rosie Campbell}, \bibinfo{person}{Andrew Cann}, \bibinfo{person}{Brittany Carey}, \bibinfo{person}{Chelsea Carlson}, \bibinfo{person}{Rory Carmichael}, \bibinfo{person}{Brooke Chan}, \bibinfo{person}{Che Chang}, \bibinfo{person}{Fotis Chantzis}, \bibinfo{person}{Derek Chen}, \bibinfo{person}{Sully Chen}, \bibinfo{person}{Ruby Chen}, \bibinfo{person}{Jason Chen}, \bibinfo{person}{Mark Chen}, \bibinfo{person}{Ben Chess}, \bibinfo{person}{Chester Cho}, \bibinfo{person}{Casey Chu}, \bibinfo{person}{Hyung~Won Chung}, \bibinfo{person}{Dave Cummings}, \bibinfo{person}{Jeremiah Currier}, \bibinfo{person}{Yunxing Dai}, \bibinfo{person}{Cory Decareaux}, \bibinfo{person}{Thomas Degry}, \bibinfo{person}{Noah Deutsch}, \bibinfo{person}{Damien Deville}, \bibinfo{person}{Arka Dhar}, \bibinfo{person}{David Dohan}, \bibinfo{person}{Steve Dowling}, \bibinfo{person}{Sheila Dunning}, \bibinfo{person}{Adrien Ecoffet}, \bibinfo{person}{Atty Eleti},
  \bibinfo{person}{Tyna Eloundou}, \bibinfo{person}{David Farhi}, \bibinfo{person}{Liam Fedus}, \bibinfo{person}{Niko Felix}, \bibinfo{person}{Simón~Posada Fishman}, \bibinfo{person}{Juston Forte}, \bibinfo{person}{Isabella Fulford}, \bibinfo{person}{Leo Gao}, \bibinfo{person}{Elie Georges}, \bibinfo{person}{Christian Gibson}, \bibinfo{person}{Vik Goel}, \bibinfo{person}{Tarun Gogineni}, \bibinfo{person}{Gabriel Goh}, \bibinfo{person}{Rapha Gontijo-Lopes}, \bibinfo{person}{Jonathan Gordon}, \bibinfo{person}{Morgan Grafstein}, \bibinfo{person}{Scott Gray}, \bibinfo{person}{Ryan Greene}, \bibinfo{person}{Joshua Gross}, \bibinfo{person}{Shixiang~Shane Gu}, \bibinfo{person}{Yufei Guo}, \bibinfo{person}{Chris Hallacy}, \bibinfo{person}{Jesse Han}, \bibinfo{person}{Jeff Harris}, \bibinfo{person}{Yuchen He}, \bibinfo{person}{Mike Heaton}, \bibinfo{person}{Johannes Heidecke}, \bibinfo{person}{Chris Hesse}, \bibinfo{person}{Alan Hickey}, \bibinfo{person}{Wade Hickey}, \bibinfo{person}{Peter Hoeschele},
  \bibinfo{person}{Brandon Houghton}, \bibinfo{person}{Kenny Hsu}, \bibinfo{person}{Shengli Hu}, \bibinfo{person}{Xin Hu}, \bibinfo{person}{Joost Huizinga}, \bibinfo{person}{Shantanu Jain}, \bibinfo{person}{Shawn Jain}, \bibinfo{person}{Joanne Jang}, \bibinfo{person}{Angela Jiang}, \bibinfo{person}{Roger Jiang}, \bibinfo{person}{Haozhun Jin}, \bibinfo{person}{Denny Jin}, \bibinfo{person}{Shino Jomoto}, \bibinfo{person}{Billie Jonn}, \bibinfo{person}{Heewoo Jun}, \bibinfo{person}{Tomer Kaftan}, \bibinfo{person}{Łukasz Kaiser}, \bibinfo{person}{Ali Kamali}, \bibinfo{person}{Ingmar Kanitscheider}, \bibinfo{person}{Nitish~Shirish Keskar}, \bibinfo{person}{Tabarak Khan}, \bibinfo{person}{Logan Kilpatrick}, \bibinfo{person}{Jong~Wook Kim}, \bibinfo{person}{Christina Kim}, \bibinfo{person}{Yongjik Kim}, \bibinfo{person}{Jan~Hendrik Kirchner}, \bibinfo{person}{Jamie Kiros}, \bibinfo{person}{Matt Knight}, \bibinfo{person}{Daniel Kokotajlo}, \bibinfo{person}{Łukasz Kondraciuk}, \bibinfo{person}{Andrew Kondrich},
  \bibinfo{person}{Aris Konstantinidis}, \bibinfo{person}{Kyle Kosic}, \bibinfo{person}{Gretchen Krueger}, \bibinfo{person}{Vishal Kuo}, \bibinfo{person}{Michael Lampe}, \bibinfo{person}{Ikai Lan}, \bibinfo{person}{Teddy Lee}, \bibinfo{person}{Jan Leike}, \bibinfo{person}{Jade Leung}, \bibinfo{person}{Daniel Levy}, \bibinfo{person}{Chak~Ming Li}, \bibinfo{person}{Rachel Lim}, \bibinfo{person}{Molly Lin}, \bibinfo{person}{Stephanie Lin}, \bibinfo{person}{Mateusz Litwin}, \bibinfo{person}{Theresa Lopez}, \bibinfo{person}{Ryan Lowe}, \bibinfo{person}{Patricia Lue}, \bibinfo{person}{Anna Makanju}, \bibinfo{person}{Kim Malfacini}, \bibinfo{person}{Sam Manning}, \bibinfo{person}{Todor Markov}, \bibinfo{person}{Yaniv Markovski}, \bibinfo{person}{Bianca Martin}, \bibinfo{person}{Katie Mayer}, \bibinfo{person}{Andrew Mayne}, \bibinfo{person}{Bob McGrew}, \bibinfo{person}{Scott~Mayer McKinney}, \bibinfo{person}{Christine McLeavey}, \bibinfo{person}{Paul McMillan}, \bibinfo{person}{Jake McNeil}, \bibinfo{person}{David
  Medina}, \bibinfo{person}{Aalok Mehta}, \bibinfo{person}{Jacob Menick}, \bibinfo{person}{Luke Metz}, \bibinfo{person}{Andrey Mishchenko}, \bibinfo{person}{Pamela Mishkin}, \bibinfo{person}{Vinnie Monaco}, \bibinfo{person}{Evan Morikawa}, \bibinfo{person}{Daniel Mossing}, \bibinfo{person}{Tong Mu}, \bibinfo{person}{Mira Murati}, \bibinfo{person}{Oleg Murk}, \bibinfo{person}{David Mély}, \bibinfo{person}{Ashvin Nair}, \bibinfo{person}{Reiichiro Nakano}, \bibinfo{person}{Rajeev Nayak}, \bibinfo{person}{Arvind Neelakantan}, \bibinfo{person}{Richard Ngo}, \bibinfo{person}{Hyeonwoo Noh}, \bibinfo{person}{Long Ouyang}, \bibinfo{person}{Cullen O'Keefe}, \bibinfo{person}{Jakub Pachocki}, \bibinfo{person}{Alex Paino}, \bibinfo{person}{Joe Palermo}, \bibinfo{person}{Ashley Pantuliano}, \bibinfo{person}{Giambattista Parascandolo}, \bibinfo{person}{Joel Parish}, \bibinfo{person}{Emy Parparita}, \bibinfo{person}{Alex Passos}, \bibinfo{person}{Mikhail Pavlov}, \bibinfo{person}{Andrew Peng}, \bibinfo{person}{Adam
  Perelman}, \bibinfo{person}{Filipe de Avila Belbute~Peres}, \bibinfo{person}{Michael Petrov}, \bibinfo{person}{Henrique~Ponde de Oliveira~Pinto}, \bibinfo{person}{Michael}, \bibinfo{person}{Pokorny}, \bibinfo{person}{Michelle Pokrass}, \bibinfo{person}{Vitchyr~H. Pong}, \bibinfo{person}{Tolly Powell}, \bibinfo{person}{Alethea Power}, \bibinfo{person}{Boris Power}, \bibinfo{person}{Elizabeth Proehl}, \bibinfo{person}{Raul Puri}, \bibinfo{person}{Alec Radford}, \bibinfo{person}{Jack Rae}, \bibinfo{person}{Aditya Ramesh}, \bibinfo{person}{Cameron Raymond}, \bibinfo{person}{Francis Real}, \bibinfo{person}{Kendra Rimbach}, \bibinfo{person}{Carl Ross}, \bibinfo{person}{Bob Rotsted}, \bibinfo{person}{Henri Roussez}, \bibinfo{person}{Nick Ryder}, \bibinfo{person}{Mario Saltarelli}, \bibinfo{person}{Ted Sanders}, \bibinfo{person}{Shibani Santurkar}, \bibinfo{person}{Girish Sastry}, \bibinfo{person}{Heather Schmidt}, \bibinfo{person}{David Schnurr}, \bibinfo{person}{John Schulman}, \bibinfo{person}{Daniel Selsam},
  \bibinfo{person}{Kyla Sheppard}, \bibinfo{person}{Toki Sherbakov}, \bibinfo{person}{Jessica Shieh}, \bibinfo{person}{Sarah Shoker}, \bibinfo{person}{Pranav Shyam}, \bibinfo{person}{Szymon Sidor}, \bibinfo{person}{Eric Sigler}, \bibinfo{person}{Maddie Simens}, \bibinfo{person}{Jordan Sitkin}, \bibinfo{person}{Katarina Slama}, \bibinfo{person}{Ian Sohl}, \bibinfo{person}{Benjamin Sokolowsky}, \bibinfo{person}{Yang Song}, \bibinfo{person}{Natalie Staudacher}, \bibinfo{person}{Felipe~Petroski Such}, \bibinfo{person}{Natalie Summers}, \bibinfo{person}{Ilya Sutskever}, \bibinfo{person}{Jie Tang}, \bibinfo{person}{Nikolas Tezak}, \bibinfo{person}{Madeleine~B. Thompson}, \bibinfo{person}{Phil Tillet}, \bibinfo{person}{Amin Tootoonchian}, \bibinfo{person}{Elizabeth Tseng}, \bibinfo{person}{Preston Tuggle}, \bibinfo{person}{Nick Turley}, \bibinfo{person}{Jerry Tworek}, \bibinfo{person}{Juan Felipe~Cerón Uribe}, \bibinfo{person}{Andrea Vallone}, \bibinfo{person}{Arun Vijayvergiya}, \bibinfo{person}{Chelsea Voss},
  \bibinfo{person}{Carroll Wainwright}, \bibinfo{person}{Justin~Jay Wang}, \bibinfo{person}{Alvin Wang}, \bibinfo{person}{Ben Wang}, \bibinfo{person}{Jonathan Ward}, \bibinfo{person}{Jason Wei}, \bibinfo{person}{CJ Weinmann}, \bibinfo{person}{Akila Welihinda}, \bibinfo{person}{Peter Welinder}, \bibinfo{person}{Jiayi Weng}, \bibinfo{person}{Lilian Weng}, \bibinfo{person}{Matt Wiethoff}, \bibinfo{person}{Dave Willner}, \bibinfo{person}{Clemens Winter}, \bibinfo{person}{Samuel Wolrich}, \bibinfo{person}{Hannah Wong}, \bibinfo{person}{Lauren Workman}, \bibinfo{person}{Sherwin Wu}, \bibinfo{person}{Jeff Wu}, \bibinfo{person}{Michael Wu}, \bibinfo{person}{Kai Xiao}, \bibinfo{person}{Tao Xu}, \bibinfo{person}{Sarah Yoo}, \bibinfo{person}{Kevin Yu}, \bibinfo{person}{Qiming Yuan}, \bibinfo{person}{Wojciech Zaremba}, \bibinfo{person}{Rowan Zellers}, \bibinfo{person}{Chong Zhang}, \bibinfo{person}{Marvin Zhang}, \bibinfo{person}{Shengjia Zhao}, \bibinfo{person}{Tianhao Zheng}, \bibinfo{person}{Juntang Zhuang},
  \bibinfo{person}{William Zhuk}, {and} \bibinfo{person}{Barret Zoph}.} \bibinfo{year}{2024}\natexlab{}.
\newblock \bibinfo{title}{GPT-4 Technical Report}.
\newblock
\showeprint[arxiv]{2303.08774}~[cs.CL]
\urldef\tempurl%
\url{https://arxiv.org/abs/2303.08774}
\showURL{%
\tempurl}


\bibitem[Qi et~al\mbox{.}(2024)]%
        {qiComputerVisionbasedHand2024}
\bibfield{author}{\bibinfo{person}{Jing Qi}, \bibinfo{person}{Li Ma}, \bibinfo{person}{Zhenchao Cui}, {and} \bibinfo{person}{Yushu Yu}.} \bibinfo{year}{2024}\natexlab{}.
\newblock \showarticletitle{Computer Vision-Based Hand Gesture Recognition for Human-Robot Interaction: A Review}.
\newblock \bibinfo{journal}{\emph{Complex \& Intelligent Systems}} \bibinfo{volume}{10}, \bibinfo{number}{1} (\bibinfo{date}{Feb.} \bibinfo{year}{2024}), \bibinfo{pages}{1581--1606}.
\newblock
\showISSN{2199-4536, 2198-6053}
\href{https://doi.org/10.1007/s40747-023-01173-6}{doi:\nolinkurl{10.1007/s40747-023-01173-6}}


\bibitem[Qi et~al\mbox{.}(2023)]%
        {10.1145/3610910}
\bibfield{author}{\bibinfo{person}{Xiangyao Qi}, \bibinfo{person}{Qi Lu}, \bibinfo{person}{Wentao Pan}, \bibinfo{person}{Yingying Zhao}, \bibinfo{person}{Rui Zhu}, \bibinfo{person}{Mingzhi Dong}, \bibinfo{person}{Yuhu Chang}, \bibinfo{person}{Qin Lv}, \bibinfo{person}{Robert~P. Dick}, \bibinfo{person}{Fan Yang}, \bibinfo{person}{Tun Lu}, \bibinfo{person}{Ning Gu}, {and} \bibinfo{person}{Li Shang}.} \bibinfo{year}{2023}\natexlab{}.
\newblock \showarticletitle{CASES: A Cognition-Aware Smart Eyewear System for Understanding How People Read}.
\newblock \bibinfo{journal}{\emph{Proc. ACM Interact. Mob. Wearable Ubiquitous Technol.}} \bibinfo{volume}{7}, \bibinfo{number}{3}, Article \bibinfo{articleno}{115} (\bibinfo{date}{Sept.} \bibinfo{year}{2023}), \bibinfo{numpages}{31}~pages.
\newblock
\href{https://doi.org/10.1145/3610910}{doi:\nolinkurl{10.1145/3610910}}


\bibitem[Rastgoo et~al\mbox{.}(2024)]%
        {rastgooMultimodalZeroshotDynamic2024}
\bibfield{author}{\bibinfo{person}{Razieh Rastgoo}, \bibinfo{person}{Kourosh Kiani}, \bibinfo{person}{Sergio Escalera}, {and} \bibinfo{person}{Mohammad Sabokrou}.} \bibinfo{year}{2024}\natexlab{}.
\newblock \showarticletitle{Multi-Modal Zero-Shot Dynamic Hand Gesture Recognition}.
\newblock \bibinfo{journal}{\emph{Expert Systems with Applications}}  \bibinfo{volume}{247} (\bibinfo{date}{Aug.} \bibinfo{year}{2024}), \bibinfo{pages}{123349}.
\newblock
\showISSN{09574174}
\showLCCN{C{\textbar}8.5{\textbar}Q1{\textbar}1区TOP}
\href{https://doi.org/10.1016/j.eswa.2024.123349}{doi:\nolinkurl{10.1016/j.eswa.2024.123349}}


\bibitem[Rodol{\`a} et~al\mbox{.}(2017)]%
        {Rodol2017SHREC1}
\bibfield{author}{\bibinfo{person}{Emanuele Rodol{\`a}}, \bibinfo{person}{Luca Cosmo}, \bibinfo{person}{Or Litany}, \bibinfo{person}{Michael~M. Bronstein}, \bibinfo{person}{Alexander~M. Bronstein}, \bibinfo{person}{N. Audebert}, \bibinfo{person}{A.~Ben Hamza}, \bibinfo{person}{Alexandre Boulch}, \bibinfo{person}{Umberto Castellani}, \bibinfo{person}{Minh~N. Do}, \bibinfo{person}{Anh~Duc Duong}, \bibinfo{person}{Andrea Gasparetto}, \bibinfo{person}{Y. Hong}, \bibinfo{person}{J. Kim}, \bibinfo{person}{B.~L. Saux}, \bibinfo{person}{Roee Litman}, \bibinfo{person}{Majid Masoumi}, \bibinfo{person}{Giorgia Minello}, \bibinfo{person}{Ryutarou Ohbuchi}, \bibinfo{person}{Thuyen~V. Phan}, \bibinfo{person}{M. Rezaei}, \bibinfo{person}{A. Torsello}, \bibinfo{person}{Minh-Triet Tran}, \bibinfo{person}{Q.~T. Tran}, \bibinfo{person}{Bao Truong}, \bibinfo{person}{Lili Wan}, {and} \bibinfo{person}{Changqing Zou}.} \bibinfo{year}{2017}\natexlab{}.
\newblock \showarticletitle{SHREC ’ 17 : Deformable Shape Retrieval with Missing Parts}.
\newblock
\urldef\tempurl%
\url{https://api.semanticscholar.org/CorpusID:3993253}
\showURL{%
\tempurl}


\bibitem[Shen et~al\mbox{.}(2024)]%
        {shenOpenWorldGestureRecognition2024}
\bibfield{author}{\bibinfo{person}{Junxiao Shen}, \bibinfo{person}{Matthias~De Lange}, \bibinfo{person}{Xuhai~"Orson" Xu}, \bibinfo{person}{Enmin Zhou}, \bibinfo{person}{Ran Tan}, \bibinfo{person}{Naveen Suda}, \bibinfo{person}{Maciej Lazarewicz}, \bibinfo{person}{Per~Ola Kristensson}, \bibinfo{person}{Amy Karlson}, {and} \bibinfo{person}{Evan Strasnick}.} \bibinfo{year}{2024}\natexlab{}.
\newblock \bibinfo{title}{Towards {{Open-World Gesture Recognition}}}.
\newblock
\showLCCN{arXiv}
\href{https://doi.org/10.48550/arXiv.2401.11144}{doi:\nolinkurl{10.48550/arXiv.2401.11144}}
\showeprint[arxiv]{2401.11144}~[cs]


\bibitem[Shimon et~al\mbox{.}(2024)]%
        {shimonExploringUnimanualEar2024}
\bibfield{author}{\bibinfo{person}{Shaikh Shawon~Arefin Shimon}, \bibinfo{person}{Ali Neshati}, \bibinfo{person}{Junwei Sun}, \bibinfo{person}{Qiang Xu}, {and} \bibinfo{person}{Jian Zhao}.} \bibinfo{year}{2024}\natexlab{}.
\newblock \showarticletitle{Exploring {{Uni-manual Around Ear Off-Device Gestures}} for {{Earables}}}.
\newblock \bibinfo{journal}{\emph{Proceedings of the ACM on Interactive, Mobile, Wearable and Ubiquitous Technologies}} \bibinfo{volume}{8}, \bibinfo{number}{1} (\bibinfo{date}{March} \bibinfo{year}{2024}), \bibinfo{pages}{1--29}.
\newblock
\showISSN{2474-9567}
\href{https://doi.org/10.1145/3643513}{doi:\nolinkurl{10.1145/3643513}}


\bibitem[Sun et~al\mbox{.}(2024a)]%
        {sunMultimodalDailyLifeLogging2024}
\bibfield{author}{\bibinfo{person}{Ke Sun}, \bibinfo{person}{Chunyu Xia}, \bibinfo{person}{Xinyu Zhang}, \bibinfo{person}{Hao Chen}, {and} \bibinfo{person}{Charlie~Jianzhong Zhang}.} \bibinfo{year}{2024}\natexlab{a}.
\newblock \showarticletitle{Multimodal {{Daily-Life Logging}} in {{Free-living Environment Using Non-Visual Egocentric Sensors}} on a {{Smartphone}}}.
\newblock \bibinfo{journal}{\emph{Proceedings of the ACM on Interactive, Mobile, Wearable and Ubiquitous Technologies}} \bibinfo{volume}{8}, \bibinfo{number}{1} (\bibinfo{date}{March} \bibinfo{year}{2024}), \bibinfo{pages}{1--32}.
\newblock
\showISSN{2474-9567}
\href{https://doi.org/10.1145/3643553}{doi:\nolinkurl{10.1145/3643553}}


\bibitem[Sun et~al\mbox{.}(2024b)]%
        {sunOptimizingGestureRecognition2024}
\bibfield{author}{\bibinfo{person}{Qi Sun}, \bibinfo{person}{Tong Zhang}, \bibinfo{person}{Shang Gao}, \bibinfo{person}{Liuqingqing Yang}, {and} \bibinfo{person}{Fenghua Shao}.} \bibinfo{year}{2024}\natexlab{b}.
\newblock \bibinfo{title}{Optimizing {{Gesture Recognition}} for {{Seamless UI Interaction Using Convolutional Neural Networks}}}.
\newblock
\href{https://doi.org/10.48550/arXiv.2411.15598}{doi:\nolinkurl{10.48550/arXiv.2411.15598}}
\showeprint[arxiv]{2411.15598}~[cs]


\bibitem[Wan et~al\mbox{.}(2022)]%
        {9172121}
\bibfield{author}{\bibinfo{person}{Jun Wan}, \bibinfo{person}{Chi Lin}, \bibinfo{person}{Longyin Wen}, \bibinfo{person}{Yunan Li}, \bibinfo{person}{Qiguang Miao}, \bibinfo{person}{Sergio Escalera}, \bibinfo{person}{Gholamreza Anbarjafari}, \bibinfo{person}{Isabelle Guyon}, \bibinfo{person}{Guodong Guo}, {and} \bibinfo{person}{Stan~Z. Li}.} \bibinfo{year}{2022}\natexlab{}.
\newblock \showarticletitle{ChaLearn Looking at People: IsoGD and ConGD Large-Scale RGB-D Gesture Recognition}.
\newblock \bibinfo{journal}{\emph{IEEE Transactions on Cybernetics}} \bibinfo{volume}{52}, \bibinfo{number}{5} (\bibinfo{year}{2022}), \bibinfo{pages}{3422--3433}.
\newblock
\href{https://doi.org/10.1109/TCYB.2020.3012092}{doi:\nolinkurl{10.1109/TCYB.2020.3012092}}


\bibitem[Wang et~al\mbox{.}(2024)]%
        {wangUFaceYourSmartphone2024}
\bibfield{author}{\bibinfo{person}{Shuning Wang}, \bibinfo{person}{Linghui Zhong}, \bibinfo{person}{Yongjian Fu}, \bibinfo{person}{Lili Chen}, \bibinfo{person}{Ju Ren}, {and} \bibinfo{person}{Yaoxue Zhang}.} \bibinfo{year}{2024}\natexlab{}.
\newblock \showarticletitle{{{UFace}}: {{Your Smartphone Can}} "{{Hear}}" {{Your Facial Expression}}!}
\newblock \bibinfo{journal}{\emph{Proceedings of the ACM on Interactive, Mobile, Wearable and Ubiquitous Technologies}} \bibinfo{volume}{8}, \bibinfo{number}{1} (\bibinfo{date}{March} \bibinfo{year}{2024}), \bibinfo{pages}{1--27}.
\newblock
\showISSN{2474-9567}
\href{https://doi.org/10.1145/3643546}{doi:\nolinkurl{10.1145/3643546}}


\bibitem[Wei et~al\mbox{.}(2022)]%
        {10.5555/3600270.3602070}
\bibfield{author}{\bibinfo{person}{Jason Wei}, \bibinfo{person}{Xuezhi Wang}, \bibinfo{person}{Dale Schuurmans}, \bibinfo{person}{Maarten Bosma}, \bibinfo{person}{Brian Ichter}, \bibinfo{person}{Fei Xia}, \bibinfo{person}{Ed~H. Chi}, \bibinfo{person}{Quoc~V. Le}, {and} \bibinfo{person}{Denny Zhou}.} \bibinfo{year}{2022}\natexlab{}.
\newblock \showarticletitle{Chain-of-thought prompting elicits reasoning in large language models}. In \bibinfo{booktitle}{\emph{Proceedings of the 36th International Conference on Neural Information Processing Systems}} (New Orleans, LA, USA) \emph{(\bibinfo{series}{NIPS '22})}. \bibinfo{publisher}{Curran Associates Inc.}, \bibinfo{address}{Red Hook, NY, USA}, Article \bibinfo{articleno}{1800}, \bibinfo{numpages}{14}~pages.
\newblock
\showISBNx{9781713871088}


\bibitem[Xiao et~al\mbox{.}(2024)]%
        {xiaoChatCamEmbracingLLMs2024}
\bibfield{author}{\bibinfo{person}{Kaijie Xiao}, \bibinfo{person}{Yi Gao}, \bibinfo{person}{Fu Li}, \bibinfo{person}{Weifeng Xu}, \bibinfo{person}{Pengzhi Chen}, {and} \bibinfo{person}{Wei Dong}.} \bibinfo{year}{2024}\natexlab{}.
\newblock \showarticletitle{{{ChatCam}}: {{Embracing LLMs}} for {{Contextual Chatting-to-Camera}} with {{Interest-Oriented Video Summarization}}}.
\newblock \bibinfo{journal}{\emph{Proceedings of the ACM on Interactive, Mobile, Wearable and Ubiquitous Technologies}} \bibinfo{volume}{8}, \bibinfo{number}{4} (\bibinfo{date}{Nov.} \bibinfo{year}{2024}), \bibinfo{pages}{1--34}.
\newblock
\showISSN{2474-9567}
\href{https://doi.org/10.1145/3699731}{doi:\nolinkurl{10.1145/3699731}}


\bibitem[Zeng et~al\mbox{.}(2024)]%
        {zengGestureGPTZeroshotInteractive2024}
\bibfield{author}{\bibinfo{person}{Xin Zeng}, \bibinfo{person}{Xiaoyu Wang}, \bibinfo{person}{Tengxiang Zhang}, \bibinfo{person}{Chun Yu}, \bibinfo{person}{Shengdong Zhao}, {and} \bibinfo{person}{Yiqiang Chen}.} \bibinfo{year}{2024}\natexlab{}.
\newblock \bibinfo{title}{{{GestureGPT}}: {{Toward Zero-shot Interactive Gesture Understanding}} and {{Grounding}} with {{Large Language Model Agents}}}.
\newblock
\showLCCN{arXiv}
\href{https://doi.org/10.48550/arXiv.2310.12821}{doi:\nolinkurl{10.48550/arXiv.2310.12821}}
\showeprint[arxiv]{2310.12821}~[cs]


\bibitem[Zhang et~al\mbox{.}(2024a)]%
        {10.1145/3675094.3678992}
\bibfield{author}{\bibinfo{person}{Dell Zhang}, \bibinfo{person}{Yongxiang Li}, \bibinfo{person}{Zhongjiang He}, {and} \bibinfo{person}{Xuelong Li}.} \bibinfo{year}{2024}\natexlab{a}.
\newblock \showarticletitle{Empowering Smart Glasses with Large Language Models: Towards Ubiquitous AGI}. In \bibinfo{booktitle}{\emph{Companion of the 2024 on ACM International Joint Conference on Pervasive and Ubiquitous Computing}} (Melbourne VIC, Australia) \emph{(\bibinfo{series}{UbiComp '24})}. \bibinfo{publisher}{Association for Computing Machinery}, \bibinfo{address}{New York, NY, USA}, \bibinfo{pages}{631–633}.
\newblock
\showISBNx{9798400710582}
\href{https://doi.org/10.1145/3675094.3678992}{doi:\nolinkurl{10.1145/3675094.3678992}}


\bibitem[Zhang et~al\mbox{.}(2018)]%
        {8299578}
\bibfield{author}{\bibinfo{person}{Yifan Zhang}, \bibinfo{person}{Congqi Cao}, \bibinfo{person}{Jian Cheng}, {and} \bibinfo{person}{Hanqing Lu}.} \bibinfo{year}{2018}\natexlab{}.
\newblock \showarticletitle{EgoGesture: A New Dataset and Benchmark for Egocentric Hand Gesture Recognition}.
\newblock \bibinfo{journal}{\emph{IEEE Transactions on Multimedia}} \bibinfo{volume}{20}, \bibinfo{number}{5} (\bibinfo{year}{2018}), \bibinfo{pages}{1038--1050}.
\newblock
\href{https://doi.org/10.1109/TMM.2018.2808769}{doi:\nolinkurl{10.1109/TMM.2018.2808769}}


\bibitem[Zhang et~al\mbox{.}(2024b)]%
        {zhang2024videoinstructiontuningsynthetic}
\bibfield{author}{\bibinfo{person}{Yuanhan Zhang}, \bibinfo{person}{Jinming Wu}, \bibinfo{person}{Wei Li}, \bibinfo{person}{Bo Li}, \bibinfo{person}{Zejun Ma}, \bibinfo{person}{Ziwei Liu}, {and} \bibinfo{person}{Chunyuan Li}.} \bibinfo{year}{2024}\natexlab{b}.
\newblock \bibinfo{title}{Video Instruction Tuning With Synthetic Data}.
\newblock
\showeprint[arxiv]{2410.02713}~[cs.CV]
\urldef\tempurl%
\url{https://arxiv.org/abs/2410.02713}
\showURL{%
\tempurl}


\end{thebibliography}

\appendix








\appendix

\section{Dataset}

\subsection{Data Subset}
\label{app:A.1}
\begin{table}[H]
\centering
\begin{tabular}{@{}m{2.3cm} m{3cm} m{3cm} m{3cm} m{5cm}@{}}
\toprule
\textbf{Gesture Class} & \textbf{Image} & \textbf{Motion Description} & \textbf{Semantic Meaning / Intention} & \textbf{CoT Reasoning (Excerpt)} \\
\midrule
Measure &
\includegraphics[width=\linewidth]{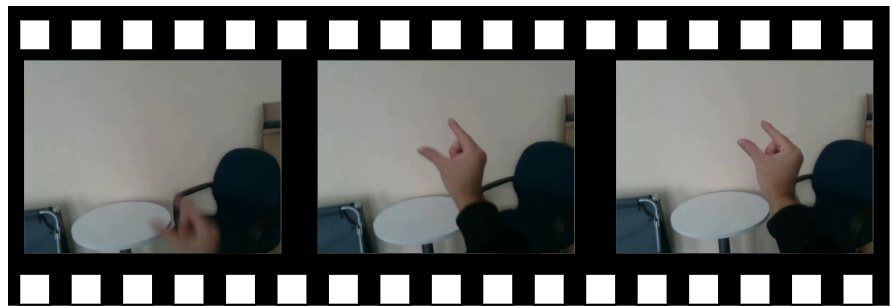}&
Extending the thumb and index finger …… &
This gesture is often
used to give a rough
idea of the length
or gap between two
points, conveying the
concept of measure-
ment through finger
positioning. &
“\textless{}think\textgreater{}……where the distance between the fingertips serves as the measurable span. ……\textless{}/think\textgreater{}\textless{}answer\textgreater{}This gesture is often used to give a rough idea of the length or gap between two points, conveying the concept of measurement through finger positioning.\textless{}/answer\textgreater{}" \\
\midrule
Phone call &
\includegraphics[width=3cm]{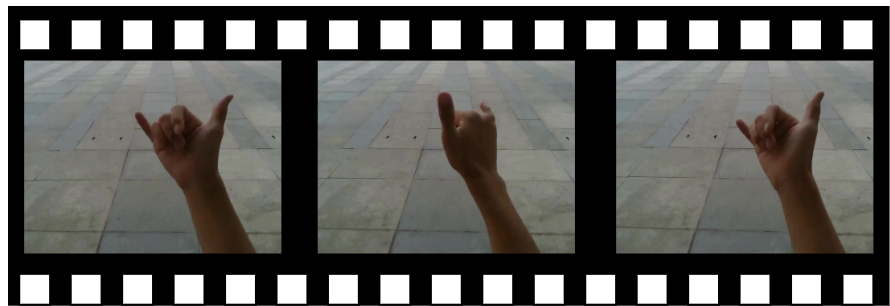} &
Extending the thumb and pinky finger while curling the other fingers inward …… &
Indicates the action of making a phone call or suggesting someone should call you. &
“\textless{}think\textgreater{}……mimics the shape and position of a traditional telephone handset. ……\textless{}/think\textgreater{}\textless{}answer\textgreater{}This gesture is used to initiate a phone call to someone.\textless{}/answer\textgreater{}" \\
\midrule
Thumb up &
\includegraphics[width=3cm]{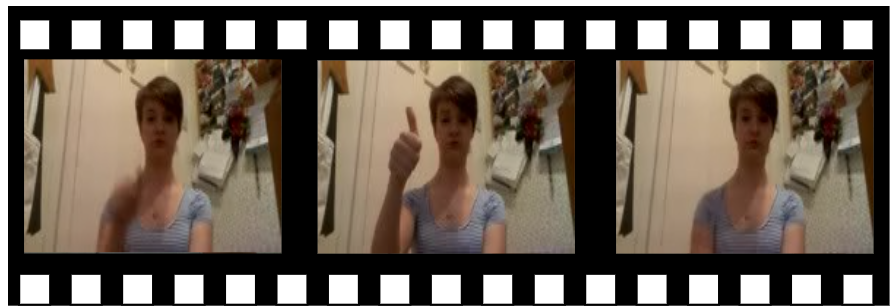} &
Raising the thumb upward while the other fingers are curled into the palm &
Represents approval, agreement, or a positive acknowledgment. Commonly used in many cultures to signal that something is good or satisfactory. &
“\textless{}think\textgreater{}……This gesture is widely recognized and utilized ……\textless{}/think\textgreater{}\textless{}answer\textgreater{}Represents approval, agreement, or a positive acknowledgment. Commonly used in many cultures to signal that something is good or satisfactory.\textless{}/answer\textgreater{}" \\
\midrule
Push hand away &
\includegraphics[width=3cm]{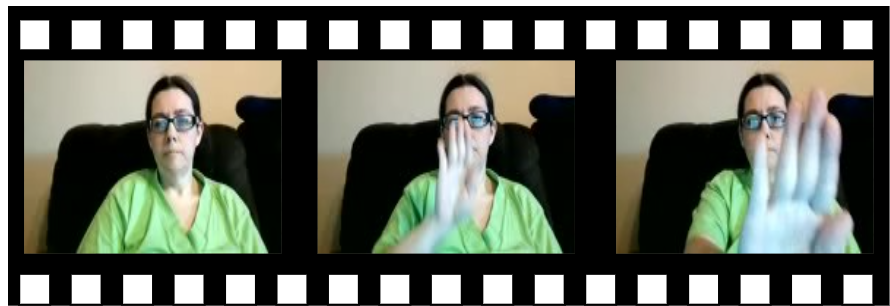} &
Extending the arm forward with the palm facing outward and fingers spread &
It can signify refusal, disapproval, or a request for personal space. This gesture signals to stop or prevents entry into a particular area. &
“\textless{}think\textgreater{}……create a physical or metaphorical boundary. ……\textless{}/think\textgreater{}\textless{}answer\textgreater{}It can signify refusal, disapproval, or a request for personal space. This gesture signals to stop or prevents entry into a particular area.\textless{}/answer\textgreater{}" \\
\bottomrule
\end{tabular}
\caption{Examples of GestureInt annotations, including image, semantic breakdown, and reasoning.}
\label{tab:gesture_examples1}
\end{table}

\clearpage

\subsection{Data Annotation}
\label{app:A.2}
First, we input the collected gesture videos and prompts that ask to caption the video with description and meaning into ChatGPT-4o. Then we have experts check whether the output from ChatGPT-4o matches the gesture category, and make modifications if there are discrepancies. This generates a dataset of captions that describe the gestures and their meaning, which we use as the training data for the first stage of training. The prompt given to GPT-4o as follow:
\\~\\
\textit{'role': 'user',\\
'content':
    'Please caption the following hand gesture video by providing a detailed description of hands and its potential meaning:\\
    Gesture Video: $\{$video\_data$\}$\\
    Provide your response only as a Python dictionary string with keys, 'description', and 'meaning'.\\
    - 'description' should be a clear, concise explanation of the gesture's physical appearance and common usage.\\
    - 'meaning' should explain the potential interpretation or cultural significance of the gesture.\\
    DO NOT PROVIDE ANY OTHER OUTPUT TEXT OR EXPLANATION. Only provide the Python dictionary string.\\
    For example, your response should look like this: $\{$'description': 'Raising the thumb upward while other fingers are curled.', 'meaning': 'Represents approval or agreement.'$\}$}'
\\~\\
For the second phase, we provide ChatGPT-4o with a caption describing a gesture and explicitly inform it of the intended meaning or purpose of the gesture, instructing ChatGPT-4o to reason through the relationship between the description of the gesture and its intended meaning. And ask it to put his thought process in $<think>\{reason\}</think>$, ending with $<answer>\{gesture\ meaning\}</answer>$. The prompt given to ChatGPT-4o as follow:
\\~\\
\textit{'role': 'user',\\
'content':
    'Given the following gesture description and its intended meaning, please explain the reasoning process that connects the physical appearance of the gesture to its intended interpretation.\\
    - Gesture Description: $\{$description$\}$\\
    - Intended Meaning: $\{$meaning$\}$\\
    Write your reasoning enclosed in \textless think\textgreater~...\textless /think\textgreater~, and conclude with the inferred gesture meaning enclosed in \textless answer\textgreater~...\textless /answer\textgreater~.\\
    Only output the reasoning and answer in the specified format. DO NOT include any other text, comments, or explanations.\\
    For example, your response should look like this:\\
    \textless think\textgreater~Raising the thumb is commonly used in many cultures to indicate positivity or agreement. Since the gesture matches this form, the intended meaning is approval.\textless /think\textgreater~\\
    \textless answer\textgreater~Represents approval or agreement.\textless /answer\textgreater~'}
\\~\\
The output generated by ChatGPT-4o, becomes the target for our training data. The input for this training data consists of two components: the video data (which provides the visual information) and the prompt "Please describe the gesture in detail and provide its meaning or intent." The model is expected to generate a coherent, structured response that reflects both the description of the gesture and its intended purpose.

\section{Ablation of freezing the LLM's weights}
\label{app:B}

Freezing the LLM during Stage 1 leads to consistent improvements across all settings (table \ref{tab:exo_ablation}, table \ref{tab:ego_ablation}), especially in egocentric and open-world scenarios. This suggests that preserving the language model's prior knowledge is crucial for robust gesture understanding.
\begin{table}[H]
\caption{Ablation analysis of Gestura with LLM unfrozen in stage 1 and Gestura under \textbf{exocentric} settings.}
\begin{tabular}{cccccccc}
\toprule
\multirow{2}{*}{Method} & \multicolumn{6}{c}{Exocentric}                        \\ \cline{3-8} 
\multicolumn{2}{c}{}               & ACC          & BLEU-1 & BLEU-2 & BLEU-3 & BLEU-4 & SPICE   \\ \hline
\multirow{2}{*}{Gestura(LLM unfrozen)}  & close& 83.01  & 51.20  & 50.92  & 49.57  & 48.76  & 0.61  \\
                                    & open  & 64.61 & 34.58  & 25.15  & 19.13  & 13.97  & 0.29  \\ \hline
\multirow{2}{*}{\textbf{Gestura}}           & close  & 84.73  & 53.94  & 51.87  & 50.61  & 49.83  & 0.63 \\
                                    & open & 65.65  & 34.88  & 25.36  & 19.37  & 14.17  & 0.30 \\ \bottomrule
\end{tabular}

\label{tab:exo_ablation}
\end{table}
\begin{table}[H]
\caption{Ablation analysis of Gestura with LLM unfrozen in stage 1 and Gestura under \textbf{egocentric} settings.}
\begin{tabular}{cccccccc}
\toprule
\multirow{2}{*}{Method} & \multicolumn{6}{c}{Egocentric}                    \\ \cline{3-8} 
\multicolumn{2}{c}{}            & ACC              & BLEU-1 & BLEU-2 & BLEU-3 & BLEU-4 & SPICE  \\ \hline
\multirow{2}{*}{Gestura(LLM unfrozen)}  & close  & 56.38& 47.03  & 41.32  & 38.21  & 36.50  & 0.48  \\
                                    & open  & 20.18  & 34.21  & 22.50  & 14.82  & 9.85   & 0.20 \\ \hline
\multirow{2}{*}{\textbf{Gestura}}           & close  & 66.14 & 52.33  & 47.72  & 45.15  & 43.67  & 0.56 \\
                                    & open & 21.71  & 33.73  & 22.10  & 14.60  & 9.93   & 0.20  \\ \bottomrule
\end{tabular}

\label{tab:ego_ablation}
\end{table}

\section{LLM as Judge}
\label{app:C}

\begin{table}[ht]
\centering
\caption{Pairwise inter-rater agreement (Cohen $\kappa$) between the four human raters.}
\label{tab:human_kappa}
\begin{tabular}{lcccc}
\toprule
 & \textbf{human1} & \textbf{human2} & \textbf{human3} & \textbf{human4}\\
\midrule
human1 & ---   & 0.906 & 0.906 & 0.775\\
human2 & 0.906 & ---   & 0.964 & 0.883\\
human3 & 0.906 & 0.964 & ---   & 0.843\\
human4 & 0.775 & 0.883 & 0.843 & ---  \\
\bottomrule
\end{tabular}
\end{table}

\begin{table}[ht]
\centering
\caption{Agreement between GPT-4o and the human majority vote.}
\label{tab:gpt_vs_human}
\begin{tabular}{lccc}
\toprule
 & Cohen $\kappa$ & MCC & MAE (0/1)\\
\midrule
GPT-4o vs.\ Human & 0.982 & 0.982 & 0.016\\
\bottomrule
\end{tabular}
\end{table}

We evaluate the reliability of GPT-4o as an automatic judge on a 200-sample subset independently labelled by four human raters. Using majority voting, we construct a human gold standard $\mathbf{H}$ and compare it against GPT-4o's binary decisions $\mathbf{G}$.

To quantify agreement, we report three complementary metrics. The mean absolute error (MAE) between GPT and human labels is $\text{MAE}(\mathbf{G}, \mathbf{H}) = 0.016$, meaning GPT-4o's decisions differ from the human consensus on only 1.6\% of cases. Cohen's $\kappa = 0.982$ and Matthews correlation coefficient (MCC) = 0.982 both indicate great agreement, consistent with prior interpretation standards. 

We also assess potential bias and instability. The average rating difference $\bar{G} - \bar{H}$ is +0.020, suggesting GPT is 2 percentage points more lenient than humans. To test robustness, we prompted GPT-4o three times using paraphrased instructions. The standard deviation of verdicts across these runs was 0.09, indicating low instability and minimal sensitivity to prompt variation.

Overall, these results confirm that GPT-4o produces judgments that are highly consistent with expert human raters, with negligible systematic bias and strong reliability across reformulations. We therefore employ GPT-4o as the primary evaluator for all large-scale experiments in this study.

\end{document}